\documentclass[journal]{IEEEtran}

\usepackage{array}
\usepackage{epsfig}
\usepackage{amsmath}
\usepackage{amssymb}
\usepackage{color}
\usepackage{url}
\usepackage{bm}
\usepackage{subfig}
\usepackage{xfrac}
\usepackage{tikz}
\usepackage{multirow}
\usepackage{enumitem}
\usepackage{xcolor,colortbl}
\usetikzlibrary{positioning,calc}
\usetikzlibrary{shapes.geometric, arrows}

\newcolumntype{C}{>{\centering\arraybackslash}p{6em}}

\def\L{{\cal L}}

\def\I{{\cal I}}

\def\H{{\cal H}}
\def\R{{\cal R}}
\def\S{{\cal S}}

\newcommand{\bu}{{\bm u}}

\newcommand{\bz}{{\bm z}}
\newcommand{\by}{{\bm y}}
\newcommand{\bx}{{\bm x}}

\definecolor{LightRed}{rgb}{1,0.8,0.8}
\newcolumntype{u}{>{\columncolor{LightRed}}c}
\definecolor{LightGreen}{rgb}{0.8,1,0.8}
\newcolumntype{v}{>{\columncolor{LightGreen}}c}
\definecolor{LightBlue}{rgb}{0.8,0.8,1}
\newcolumntype{x}{>{\columncolor{LightBlue}}c}

\newcommand{\pLen}{0.45cm}

\tikzstyle{arrow} = [thick,->,>=stealth]

\begin{document}
%
\title{Scalable Kernel-Based Minimum Mean Square Error Estimator for Accelerated Image Error Concealment}
%
%
%

\author{J\'{a}n~Koloda, J\"{u}rgen~Seiler, Antonio~M.~Peinado, and~Andr\'{e}~Kaup
\thanks{J. Koloda, J. Seiler and A. Kaup are with the Chair of Multimedia Communications and Signal Processing, Friedrich-Alexander University (FAU) of Erlangen-Nuremberg, Erlangen, Germany.}
\thanks{A.M. Peinado is with the Department of Signal Theory, Networking and Communications, University of Granada, Granada, Spain.}
\thanks{This work has been partially supported by the Research Training Group 1773 ''Heterogeneous Image Systems'', funded by the German Research Foundation (DFG), and by the Spanish MICINN/MINECO/FEDER project TEC2013-46690-P.}
}

\maketitle

\begin{abstract}
Error concealment is of great importance for block-based video systems, such as DVB or video streaming services. In this paper, we propose a novel scalable spatial error concealment algorithm that aims at obtaining high quality reconstructions with reduced computational burden. The proposed technique exploits the excellent reconstructing abilities of the kernel-based minimum mean square error \mbox{(K-MMSE)} estimator. We propose to decompose this approach into a set of hierarchically stacked layers. The first layer performs the basic reconstruction that the subsequent layers can eventually refine. In addition, we design a layer management mechanism, based on profiles, that dynamically adapts the use of higher layers to the visual complexity of the area being reconstructed. The proposed technique outperforms other state-of-the-art algorithms and produces high quality reconstructions, equivalent to K-MMSE, while requiring around one tenth of its computational time.
\end{abstract}


\section{Introduction}

\IEEEPARstart{T}{he} transmission of video signals through various channels is steadily growing as a response to the demand of the users. Various digital video broadcasting (DVB) standards have been developed to allow for digital broadcasting over cable, satellites or terrestrial channels \cite{ISEC_multibroadcast}. In addition, the demand for video streaming services has been rapidly growing in recent years and the video content nowadays accounts for over 70\% of overall internet downstream traffic \cite{Sandvine}.

The aforementioned digital video broadcasting and streaming services rely on modern block-based video coding standards such as H.264/AVC and H.265/HEVC. High compression ratios are achieved by splitting the video frames into non-overlapping blocks that are encoded using intra- or interframe prediction \cite{Richardson2010,HEVC_overview}.  This efficient compression also involves severe distortions when data losses occur \cite{EC_DTV}. Although the aforementioned standards have introduced several error resilience tools, such as partitioning the coded bitstream into network abstraction layer (NAL) units or applying flexible macroblock ordering \cite{Richardson2010,HEVC_syntax}, achieving high quality reception is a challenging task since data streams are usually transmitted over error-prone channels.

Error concealment (EC) techniques constitute a very challenging field, since the quality of service is of utmost importance for the users \cite{FadingEC}. In many cases, retransmission of lost data is not possible due to real-time constraints or bandwidth restrictions. Both issues also impede additional transmission of media-specific forward error correction (FEC) \cite{FEC_Peinado}. Thus, error concealment (EC) methods, which are carried out at the decoder, have to be employed to reconstruct the missing pixels. In this paper, we will focus  on spatial EC (SEC) techniques that rely on the information provided by the current frame. Even though temporal information, provided by adjacent frames, can be exploited as well when concealing the errors \cite{Hybrid_EC}, SEC algorithms are necessary when all the available temporal information belongs to a different scene or there is no temporal information available at all. Furthermore, the inter-prediction scheme \cite{AVC} involves that every frame in the video sequence usually serves as a reference for, at least, one intercoded frame. Thus, high quality SEC is required since any reconstruction error will be propagated until the next undamaged intracoded frame arrives and resets the prediction error.

Various SEC algorithms have been proposed in the literature for block-coded video/images. The most fundamental techniques are based on interpolation, trying to exploit the correlations between adjacent pixels. Typically, they tend to produce moderate reconstruction errors at the expense of blurring. In \cite{BLI}, a simple spatial interpolation is used. Restoration of broken edges based on directional extrapolation was introduced in \cite{EXT}. This approach exploits the fact that high frequencies, such as edges, are visually the most relevant features. A technique including edge detectors combined with a Hough transform, a powerful tool for edge description, was utilized in \cite{Robie}. A more advanced Hough transform based method was proposed in \cite{Gharavi}. However, the performance of all these methods drops when multiple edges or fine textures are involved. Modelling natural images as Markov random fields for EC was treated in \cite{MRF}. This scheme produces relatively small squared reconstruction errors at the expense of blurring. Inpainting-based methods can also be adopted for SEC purposes \cite{Harrison,Criminisi}. Sequential pixel-wise recovery based on orientation adaptive interpolation is treated in \cite{OAI}. However, pixel by pixel recovery usually suffers from smoothing high frequency textures. Spatial EC based on edge analysis and visual clearness of the surrounding area was proposed in \cite{EVC}. In \cite{BayesianPyramid}, Bayesian restoration is combined with DCT pyramid decomposition in order to carry out a multi-scale estimation. Bilateral filtering that exploits a pair of Gaussian kernels is treated in \cite{BLF}. It produces moderate reconstruction errors but often incur blocking and blurring. Natural images are typically modelled as autoregressive (AR) processes which is a direct consequence of the high spatial pixel correlation. To determine the parameters of the AR predictor a sparsity-based approach was introduced in \cite{SLPE}. An adaptive linear predictor based on Bayesian information criterion is treated in \cite{Liu2015}. Finally, various SEC techniques in transform domains \cite{POCS, CVFSE, FOFSE} have been introduced.

Another important class of SEC techniques are the switching reconstruction algorithms. They rely on extracting various features from the available neighbouring area in order to switch among different SEC approaches. The objective is to apply the best fitting SEC approach according to the visual properties of the available surrounding region. Most of these switching SEC techniques base their decision mechanism on spatial gradient analysis, i.e., on studying the edge information \cite{MDI,Efficient_STEC,FE_SEC}. In \cite{CoopGame}, the switching is based on minimising the smoothness between the outer and the inner boundary of the missing block. The authors in \cite{KungKimKuo} combined edge recovery and selective directional interpolation in order to achieve a more visually pleasing texture reconstruction. In \cite{Hsia2016}, various interpolation approaches are combined based on a block classification. A content adaptive algorithm was introduced in \cite{CAD}. A simple interpolation is applied if there are only a few edges crossing the missing macroblock and a best-match approach is applied if the macroblock is decided to contain texture. For this algorithm, and in general for all switching SEC techniques, a correct macroblock classification is critical since an erroneous decision could have a very negative effect on the resulting quality.

In our previous work, we have proposed the kernel-based minimum mean square error (K-MMSE) estimator \cite{KMMSE}. It is based on an MMSE estimation scheme where the required probabilities have been obtained through kernel density estimation (KDE). Although K-MMSE considerably outperforms other state-of-the-art techniques, both on objectives and subjective levels \cite{KMMSE}, its main drawback is the associated computational burden. In this paper, we propose a scalable scheme for accelerated K-MMSE estimation. Given the generic \mbox{K-MMSE} framework, it can be decomposed into a set of reconstruction layers ranging from the basic layer that produces a rough yet fast estimation to more complex layers yielding high quality results. Unlike the classic switching techniques that select different reconstruction methods depending on the surrounding area \cite{CAD}, the proposed approach is carried out in a scalable way, i.e., during the concealment the lower layers feed information to higher layers. We design an adaptive layer management mechanism that selects the suitable layer based on how complex the surrounding area is. This selection mechanism is controlled by the so called profiles that adjust the trade-off between computational time and reconstruction quality. Simulations reveal that our proposal reduces the computational time by a factor  of 10 with respect to the original K-MMSE with negligible loss in reconstruction quality.

The paper is organised as follows. The reconstruction framework is detailed in Section \ref{sec:framework}. Section \ref{sec:kmmse} provides an overview of the K-MMSE estimator. The proposed scalable kernel-based estimator and its different reconstruction layers are described in Section \ref{sec:sk-kmmse}. The design of the layer selection mechanism based on reconstruction profiles is introduced in Section \ref{sec:profiles}. Simulation results are discussed in Section \ref{sec:results}. The last section is devoted to conclusions.

\section{Reconstruction framework}
\label{sec:framework}
\begin{figure}[t]
    \subfloat[\label{subfig-a:grid}]{
      \includegraphics[width=0.6\linewidth]{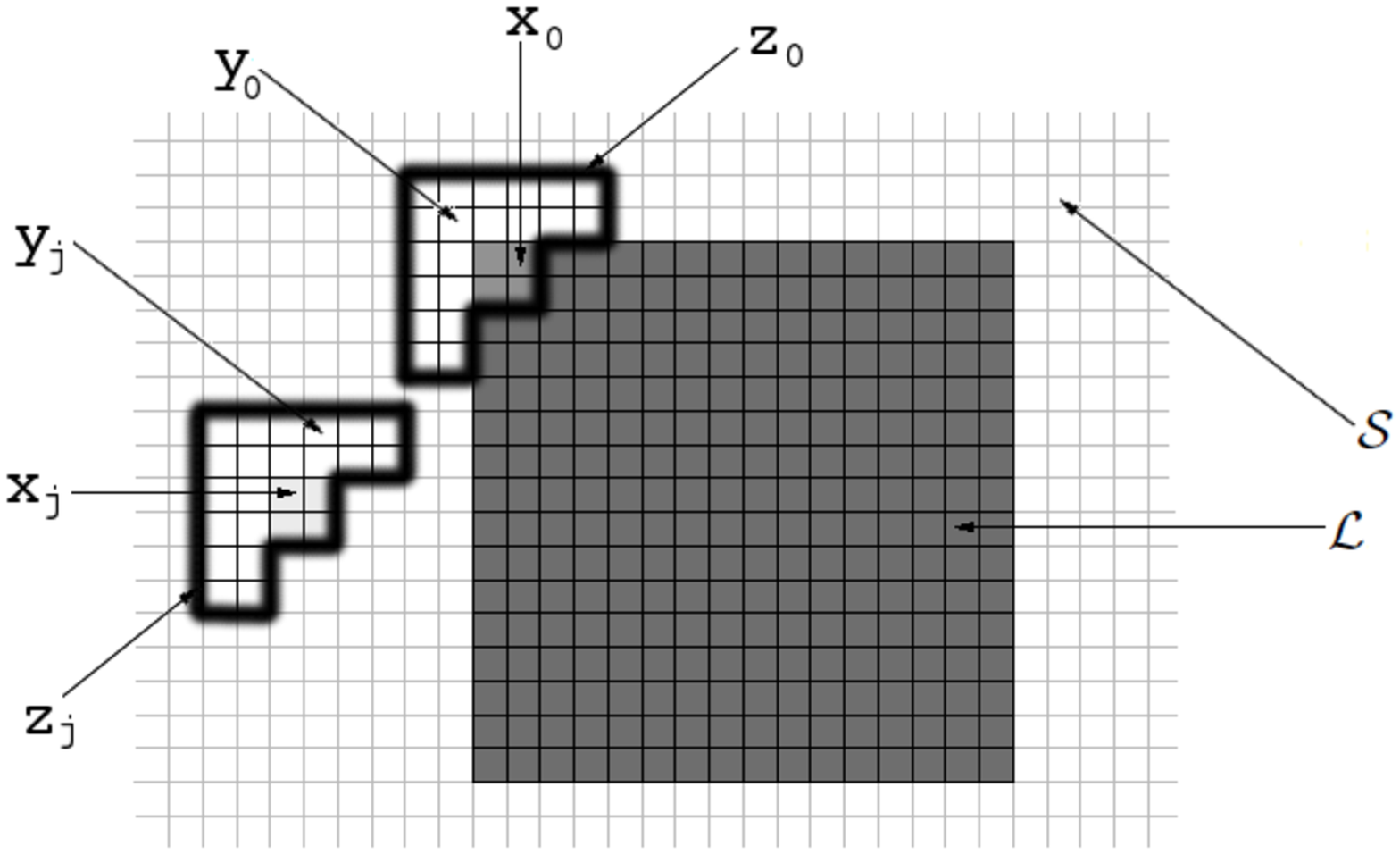}
    }
    \hfill
    \subfloat[\label{subfig-b:grid}]{%
      \includegraphics[width=0.35\linewidth]{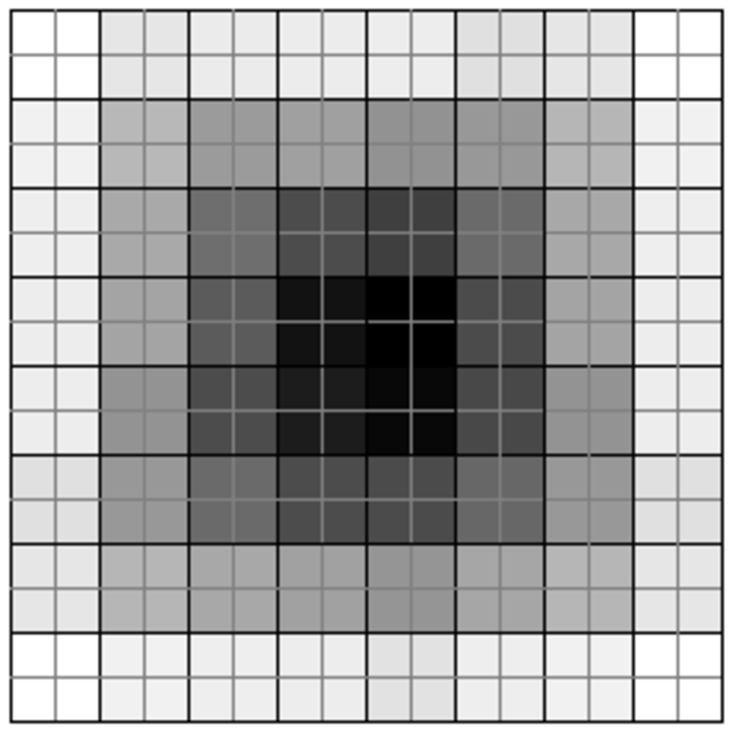}
    }
    \caption{\small{(a) Example of configuration for the vector $\bx$, $\by$ and $\bz$. (b) Filling order for sequential reconstruction using 2$\times$2 patches. The regions illustrated by brighter level are recovered first.}}
    \label{fig:grid}
\end{figure}

The reconstruction framework used in this paper will be the same as in \cite{KMMSE}. In the following, we briefly summarise it. Let $\R$ be a set of adjacent pixels that contains both missing and available samples. Let $\L$ and $\S$ denote the sets of missing and available pixels, respectively, so that $R = \L \cup \S$.
Our objective is to estimate a vector $\bx_0 \subset \L$ of missing samples. We consider that $\bx_0$ forms part of a larger vector $\bz_0 = (\bx_0^t, \by_0^t)^t$, where $\by_0 \subset \S$ is a context vector comprised by available pixels adjacent to $\bx_0$. Additionally, we will also consider all the available vectors $\bz_j = (\bx_j^t, \by_j^t)^t$ that can be built in $\S$ with the same spatial configuration as $\bz_0$. An example of such a configuration is illustrated in Fig. \ref{fig:grid}(a). In the following, $\bx_j$ and $\by_j$ will be referred to as prototype vectors and context vectors, respectively.
In our scheme, we will assume that the missing subvector $\bx_0$ is a 2$\times$2 patch of pixels and its corresponding context $\by_0$ contains all the available samples within the 6$\times$6 pixel neighbourhood centred at $\bx_0$ \cite{KMMSE}.

For testing and simulation purposes, block-based codecs employing 16$\times$16 blocks are considered. The support area $\S$ is assumed to be comprised by the 8 spatially adjacent blocks. We will apply a dispersed loss pattern, as shown in Fig. \ref{fig:example_dispersed}(b), and a random loss pattern, as shown in Fig. \ref{fig:example_random}(b). Both patterns correspond to a loss rate of approximately 25\%. Nevertheless, the proposed technique can be straightforwardly extended to any arbitrarily shaped loss pattern. Finally, the missing area $\L$ is reconstructed sequentially from its outer border towards the centre. The filling order is determined by a reliability parameter as detailed in \cite{KMMSE}. An example of this sequential filling order is illustrated in Fig. \ref{fig:grid}(b).

\section{K-MMSE overview}
\label{sec:kmmse}
In this section, the K-MMSE estimator \cite{KMMSE} is briefly summarised.
The objective is to obtain an MMSE estimate $E[\bx | \by_0]$ of the missing vector $\bx_0$ given its context $\by_0$.
We will consider the random vector variable $\bz = (\bx^t, \by^t)^t$ corresponding to the spatial configuration defined by $\bx_0$ and $\by_0$.
If the probability density function (pdf) associated to $\bz$ is available, then the desired MMSE estimate can be obtained from it. K-MMSE exploits kernel density estimation (KDE) which provides an estimate of the pdf given a set of observations $\{\bz_j; j = 1,...,M\}$ in a non parametric way, that is, avoiding any assumptions about the original pdf. This estimation is carried out by replicating and summing a kernel function at the observed vector samples. Assuming a Gaussian kernel, the pdf estimate can be expressed as

\begin{equation}
 p(\bz) = \frac{1}{M} \sum_{j=1}^M K_Z^{(j)}(\bz)
 \label{pdf}
\end{equation}
where $K_Z^{(j)}(\bz)$ is a multivariate Gaussian with mean $\bz_j$ and covariance $\H$ which is commonly referred to as bandwidth. This bandwidth matrix can be further decomposed as

\begin{equation}
 \H = 
 \begin{pmatrix}
  \H_{XX} & \H_{XY} \\
  \H_{YX} & \H_{YY}    
 \end{pmatrix}.
 \label{bandwidth}
\end{equation}
The knowledge of $p(\bz)$ allows the application of Bayesian techniques and, in particular, MMSE estimation.
The \mbox{K-MMSE} estimator can be compactly expressed as

\begin{equation}
 \hat{\bx}_0 = \tilde{\bx}_0 + \H_{XY} \H_{YY}^{-1} (\by_0 - \tilde{\by}_0)
 \label{kmmse}
\end{equation}
where $\tilde{\bx}_0$ and $\tilde{\by}_0$ are vectors linearly predicted from the sets of prototype and context vectors, respectively, that is,
\begin{equation}
  \tilde{\bx}_0 = \sum_{j=1}^M w_j(\bx_0)  
 \label{x0tilde}
\end{equation}
\begin{equation}
  \tilde{\by}_0 = \sum_{j=1}^M w_j(\by_0)  
 \label{y0tilde}
\end{equation}
For a multivariate Gaussian kernel the prediction weights $w_j$ are computed as

\begin{equation}
 w_j(\by_0) = \displaystyle{\frac{\exp\left( -\frac{1}{2} (\by_0 - \by_j)^t \H_{YY}^{-1} (\by_0 - \by_j) \right)}
	      {\sum_{i=1}^M \exp\left( -\frac{1}{2} (\by_0 - \by_i)^t \H_{YY}^{-1} (\by_0 - \by_i) \right)}}.
 \label{weights}
\end{equation}
Note that the second term of (\ref{kmmse}) is the correction vector where the unpredictable part of $\by_0$ is transformed into subspace $\bx$.

The main issue of kernel-based reconstruction is the estimation of a suitable bandwidth matrix. In \cite{KMMSE}, we have proposed a covariance submatrix scaling procedure especially designed for signal reconstruction, i.e.,

\begin{equation}
 \H = 
 \begin{pmatrix}
  \beta_{XX} C_{XX} & \beta_{XY} C_{XY} \\
  \beta_{YX} C_{YX} & \beta_{YY} C_{YY}    
 \end{pmatrix}
 \label{bandwidthScaled}
\end{equation}
where $C_{ZZ}$ is the sample covariance matrix estimated from $\bz_j (j = 1, ..., M)$ and it can be decomposed in the same way as the bandwidth matrix $\H$ in (\ref{bandwidth}), i.e.,

\begin{equation}
 C_{ZZ} = 
 \begin{pmatrix}
  C_{XX} & C_{XY} \\
  C_{YX} & C_{YY}    
 \end{pmatrix}.
 \label{covariance}
\end{equation}

Inserting (\ref{bandwidthScaled}) into (\ref{weights}) we obtain that $w_j = w_j(\beta_{YY})$. For the sake of simplicity, let us denote $\beta_{YY} = \beta$ so that $\tilde{\bx}_0 =  \tilde{\bx}_0(\beta)$ and $\tilde{\by}_0 =  \tilde{\by}_0(\beta)$. Moreover, inserting (\ref{bandwidthScaled}) into (\ref{kmmse}) we obtain that $\hat{\bx}_0 = \hat{\bx}_0(\alpha, \beta)$, where $\alpha = \sfrac{\beta_{XY}}{\beta_{YY}}$. It thus follows that this submatrix scaling scheme reduces the bandwidth estimation problem to the estimation of two parameters, $\alpha$ and $\beta$. The scale parameter $\beta$ is estimated by minimising the prediction square error over the available context $\by_0$, i.e.

\begin{equation}
 \epsilon_{\by}(\beta) = \| \by_0 - \tilde{\by}_0 \|^2 = \left\| \by_0 - \sum_{j=1}^M w_j(\by_0, \beta) \by_j \right\|^2
 \label{epsy}
\end{equation}
and the scale parameter $\beta$ is computed as
\begin{equation}
  \beta = \underset{\beta}{\operatorname{argmin}} \left( \epsilon_{\by} \right).
\end{equation}
On the other hand, since $\alpha$ controls the contribution of the correction vector in (\ref{kmmse}), it is estimated by assuming that observations in the vicinity of $\by_0$ should get similar corrections. Therefore, $\alpha$ is computed by minimising the following square error

\begin{equation}
 \epsilon_{\bx}(\alpha) = \displaystyle{\frac{1}{|\I_0|} \sum_{i \in \I_0} \|\bx_i - \hat{\bx}_i \|^2}
 \label{epsx}
\end{equation}
where $\I_0$ denotes the set of indices of observations close to $\by_0$. It has been deduced in \cite{KMMSE} that the best performance is obtained when using the $(N_y + 1)$ closest observation to $\by_0$, where $N_y$ is the dimensionality of $\by_0$. The scale parameter $\alpha$ is then computed as
\begin{equation}
  \alpha = \underset{\alpha}{\operatorname{argmin}} \left( \epsilon_{\bx} \right).
\end{equation}

Finally, after reconstructing the missing patch $\bx_0$, it is transferred to the support area $\S$ and the concealment process continues until there are no missing pixels left.

\section{Scalable K-MMSE estimator}
\label{sec:sk-kmmse}
K-MMSE is a powerful reconstruction tool that considerably outperforms other state-of-the-art reconstruction techniques especially in cases where little relevant information is available \cite{KMMSE}. However, when dealing with relatively simple structures, applying K-MMSE would be an overkill since similar reconstruction quality could be achieved by simpler (and, therefore, faster) estimators. In this section, we propose a new scalable K-MMSE estimator (SK-MMSE) that aims at accelerating the reconstruction process by decomposing the EC procedure into different reconstruction layers. The reconstruction layers obey the scalability principle so higher layers depend on the information forwarded by lower layers. Moreover, the higher the layer within the scalable EC hierarchy, the better the reconstruction at the expense of computational complexity. In the same time, SK-MMSE pursues to maintain the high reconstruction quality of the original K-MMSE algorithm. We take advantage of the fact that K-MMSE is susceptible to various reductions. The proposed SK-MMSE technique incorporates three reconstruction layers that emanate from the original K-MMSE reconstruction framework and they are introduced in the next subsections. The layer selection mechanism, controlled by profiles, is proposed in Section \ref{sec:profiles}. 

\subsection{High quality layer (HQL)}
\label{subsec:hql}
We propose the SK-MMSE algorithm to be comprised by three reconstruction layers. The highest layer, able to reconstruct complicated structures and fine textures, corresponds to the full-featured \mbox{K-MMSE} (as described in Section \ref{sec:kmmse}) and it will be referred to as high quality layer (HQL). The HQL reconstruction will therefore be equivalent to the original \mbox{K-MMSE} estimation, i.e.

\begin{equation}
  \hat{\bx}_0^{\text{HQL}} = \tilde{\bx}_0(\beta) + \alpha C_{XY}C_{YY}^{-1} \left( \by_0 - \tilde{\by}_0(\beta) \right).
\end{equation}
where the superscript refers to the corresponding reconstruction layer.

\subsection{Intermediate dynamic layer (IDL)}
\label{subsec:idl}
In many cases, we are dealing with stationary image areas, such as periodic textures or regular structures. This involves that the unpredictable part of $\by_0$ (second term of (\ref{kmmse})) can be neglected so no correction vector needs to be computed. This can be achieved by reducing the bandwidth matrix to a scalar. Thus, the correction term disappears since the corresponding submatrix $\H_{XY}$ is zero. This scheme can be further simplified by suppressing the $\beta$-optimisation procedure and computing the bandwidth as
\begin{equation}
 \H_{YY} = \sigma^2 N_y I
 \label{bw_slpe}
\end{equation}
where $I$ is the identity matrix, $\sigma^2$ is a constant factor that fixes the aperture of the Gaussian kernel and $N_y$ is introduced in order to avoid the penalisation of contexts with larger dimensionality. In other words, the bandwidth in (\ref{bw_slpe}) expresses the average distortion per context pixel. This approach is equivalent to the adaptive sparse linear predictor using exponential weights (SLP-E) as described in \cite{SLPE}. In fact, \mbox{SLP-E} can be alternatively seen as a K-MMSE estimator where the bandwidth adopts a scalar form, so no sample covariance matrix needs to be computed and, therefore, no $\alpha$-optimisation is to be performed. In addition, no $\beta$-optimisation is to be performed either since the dependency of the bandwidth with the size of the context is compensated by $N_y$. Thus, the SLP-E procedure can be interpreted as a reduction of the full-featured K-MMSE (HQL).

\begin{figure}[t]
  \begin{center}
   \begin{tikzpicture}      
      \node (support_area) {\includegraphics[width=0.5\linewidth]{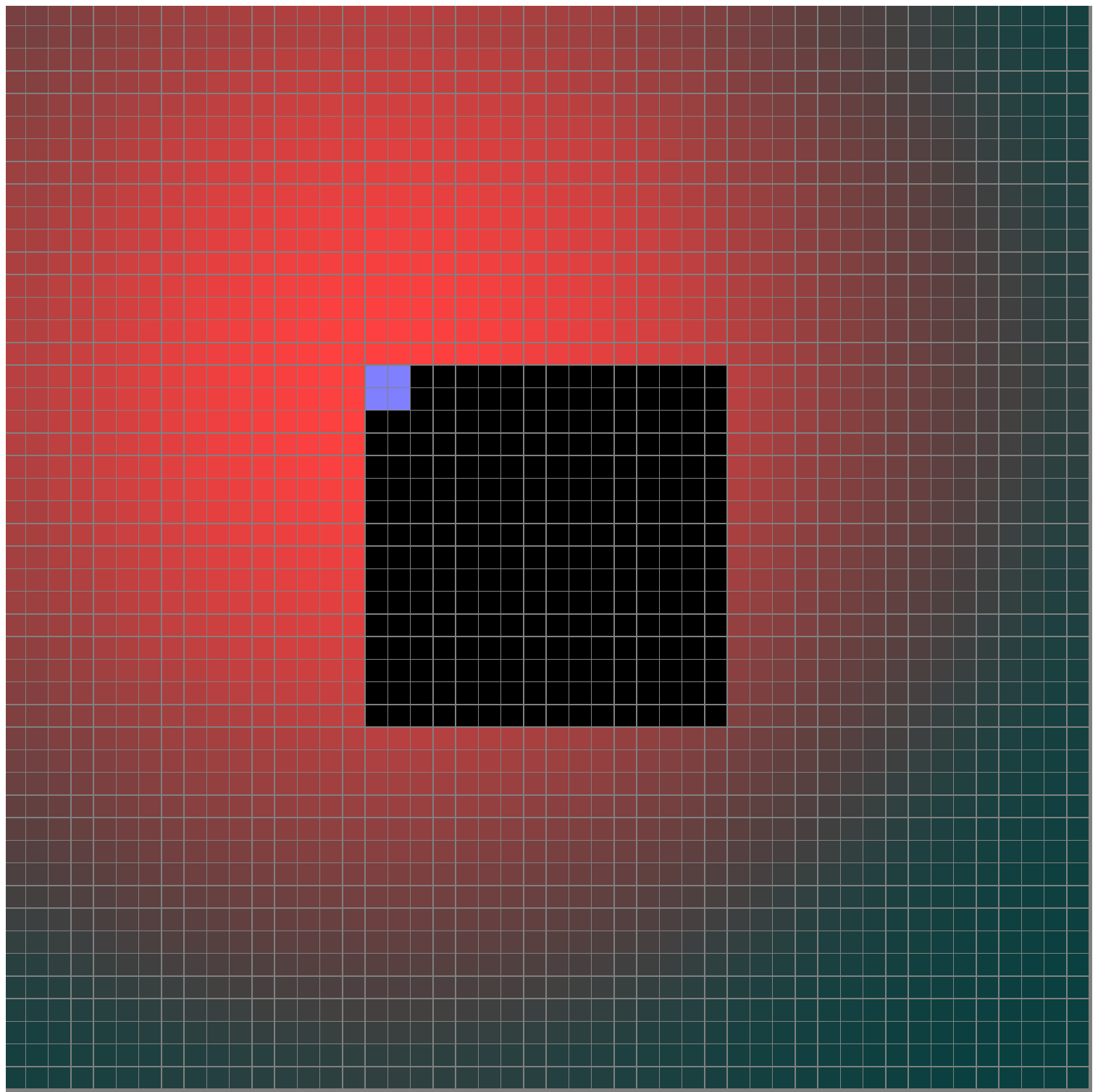}};
      
      \node (support_area_4) [right =  1.25cm of support_area] {\includegraphics[width=0.1\linewidth]{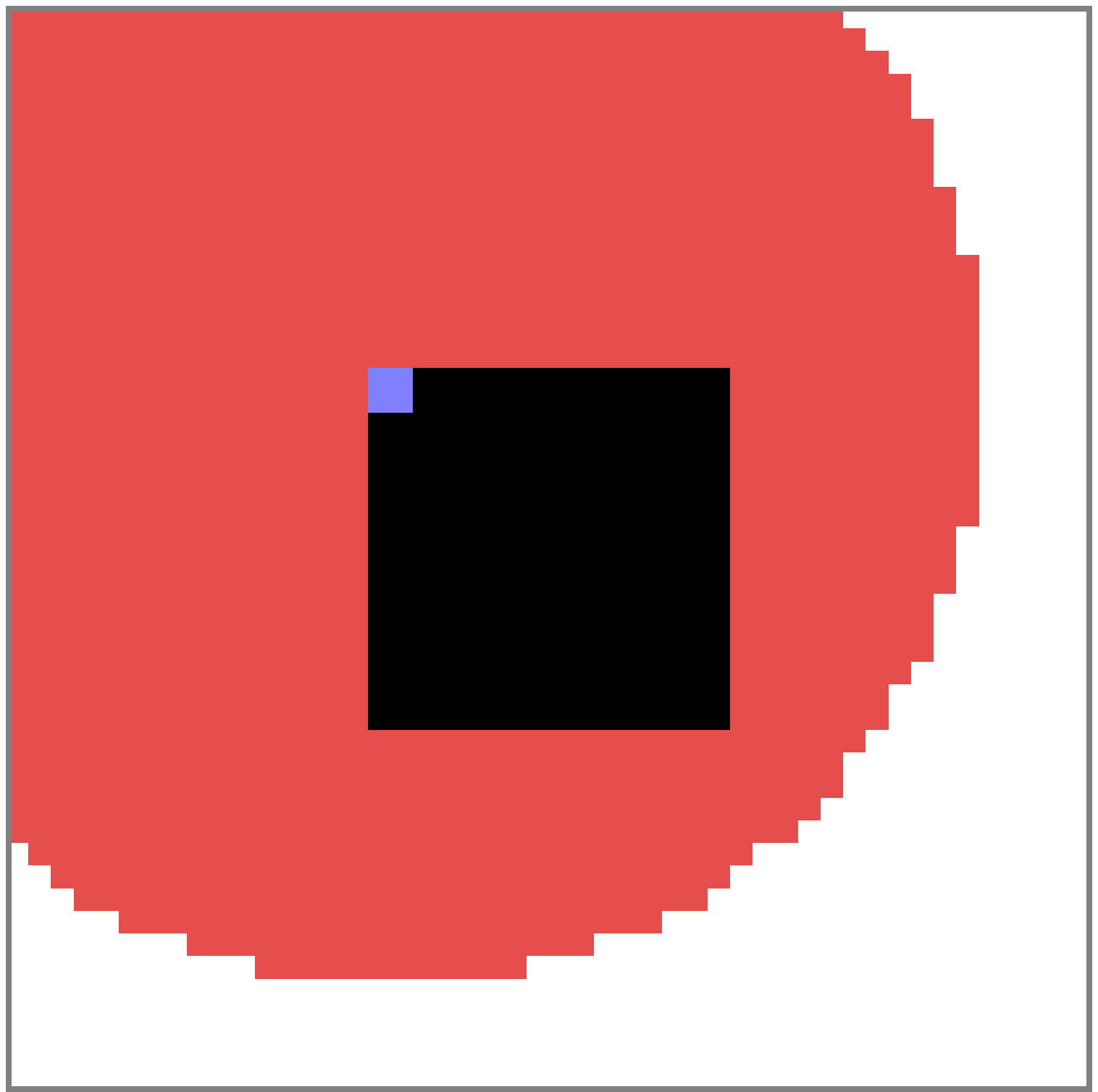}};
      \node (support_area_3) [right = 0.5cm of support_area_4] {\includegraphics[width=0.1\linewidth]{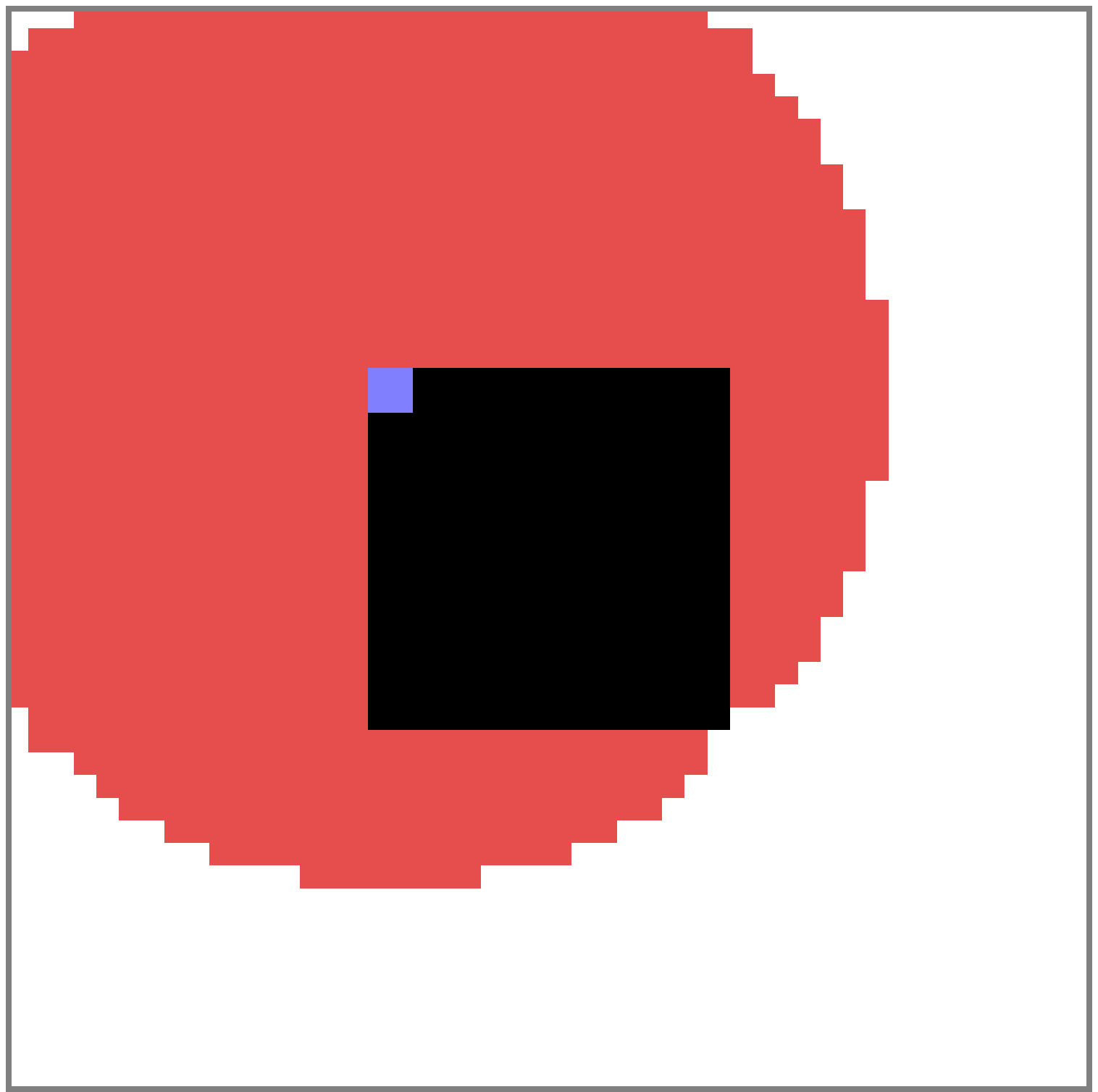}};
      
      \node (support_area_1) [above = 0.5cm of support_area_4] {\includegraphics[width=0.1\linewidth]{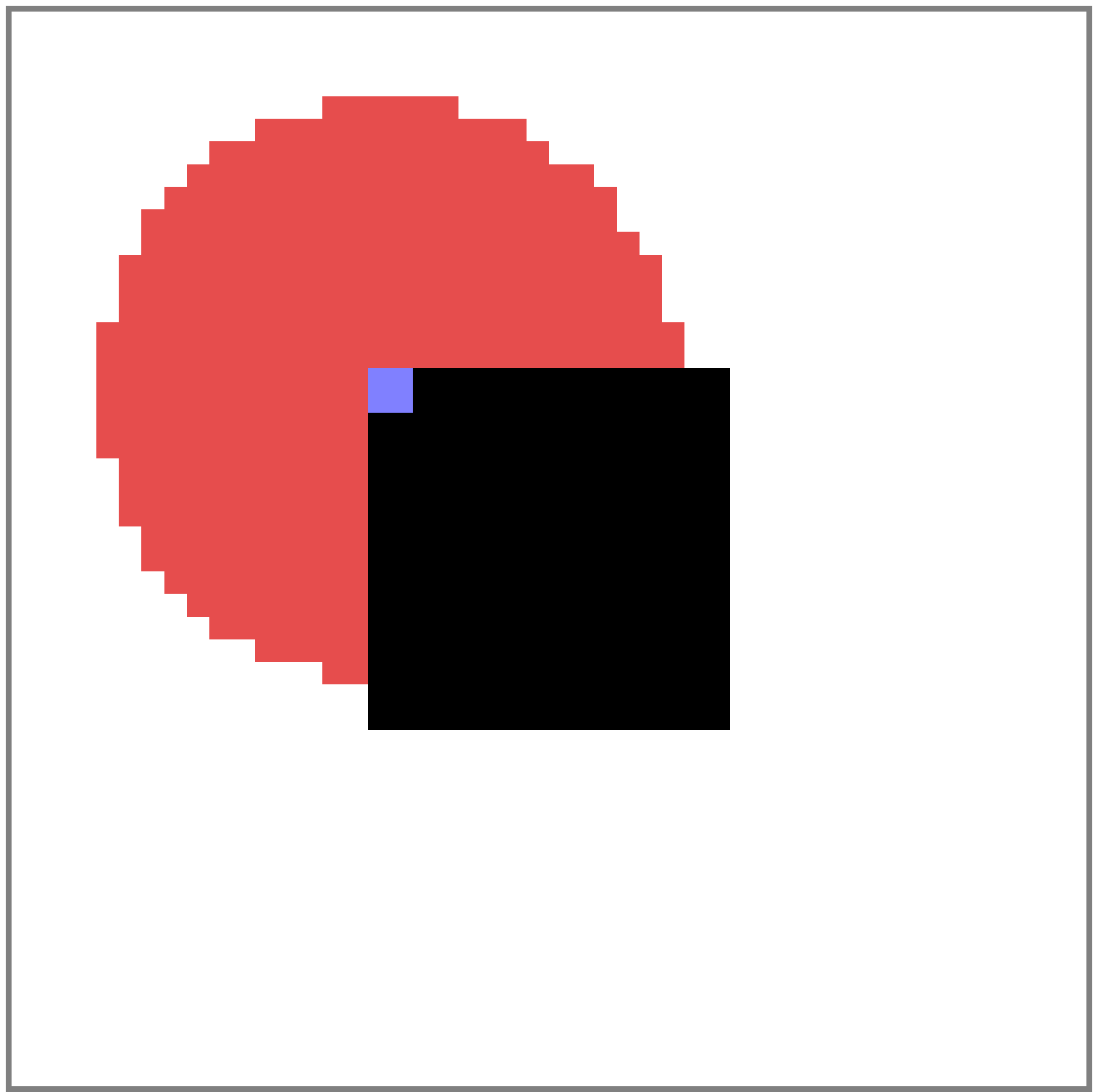}};
      \node (support_area_2) [above = 0.5cm of support_area_3] {\includegraphics[width=0.1\linewidth]{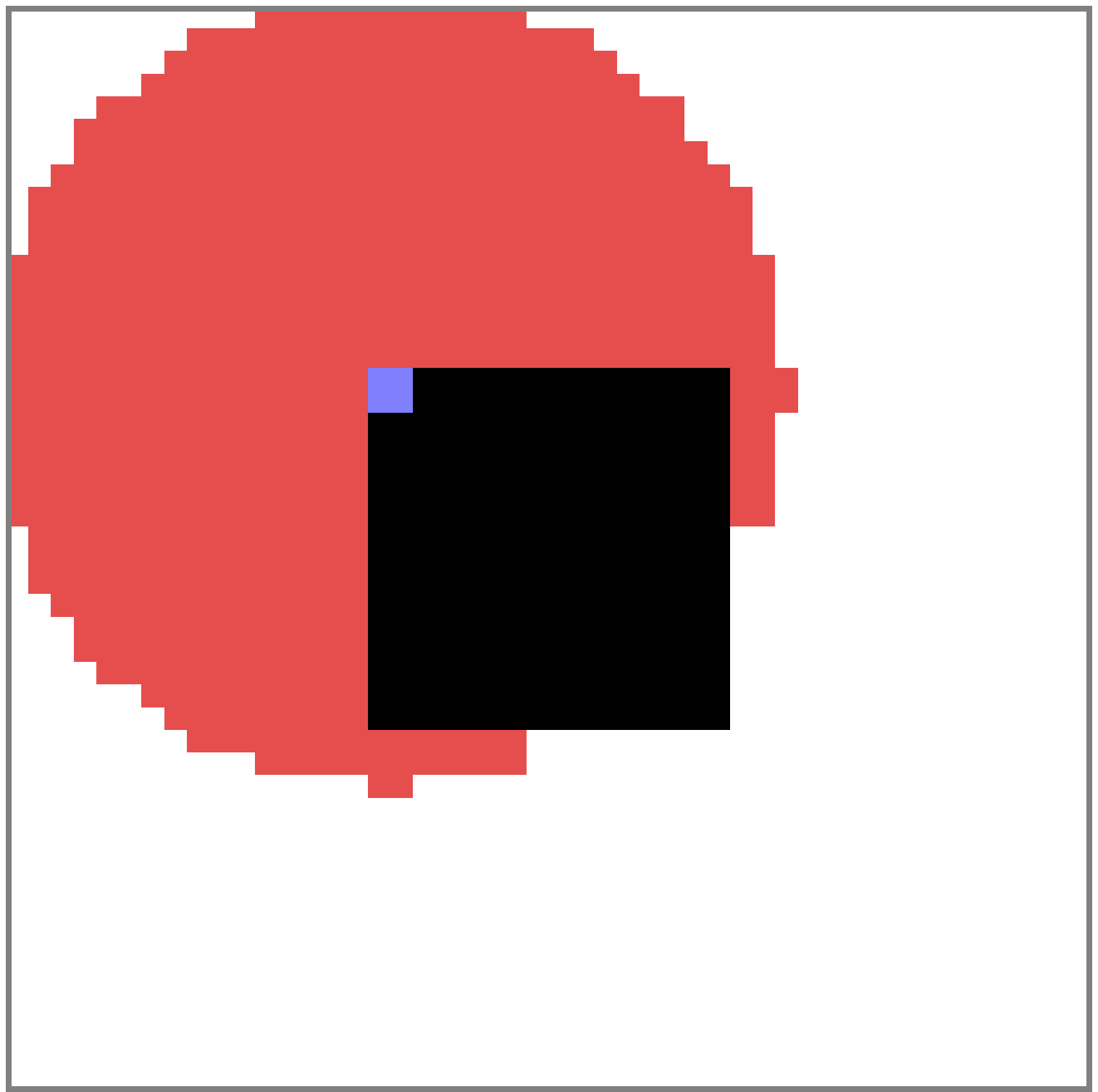}};
      
      \node (support_area_5) [below = 0.5cm of support_area_4] {\includegraphics[width=0.1\linewidth]{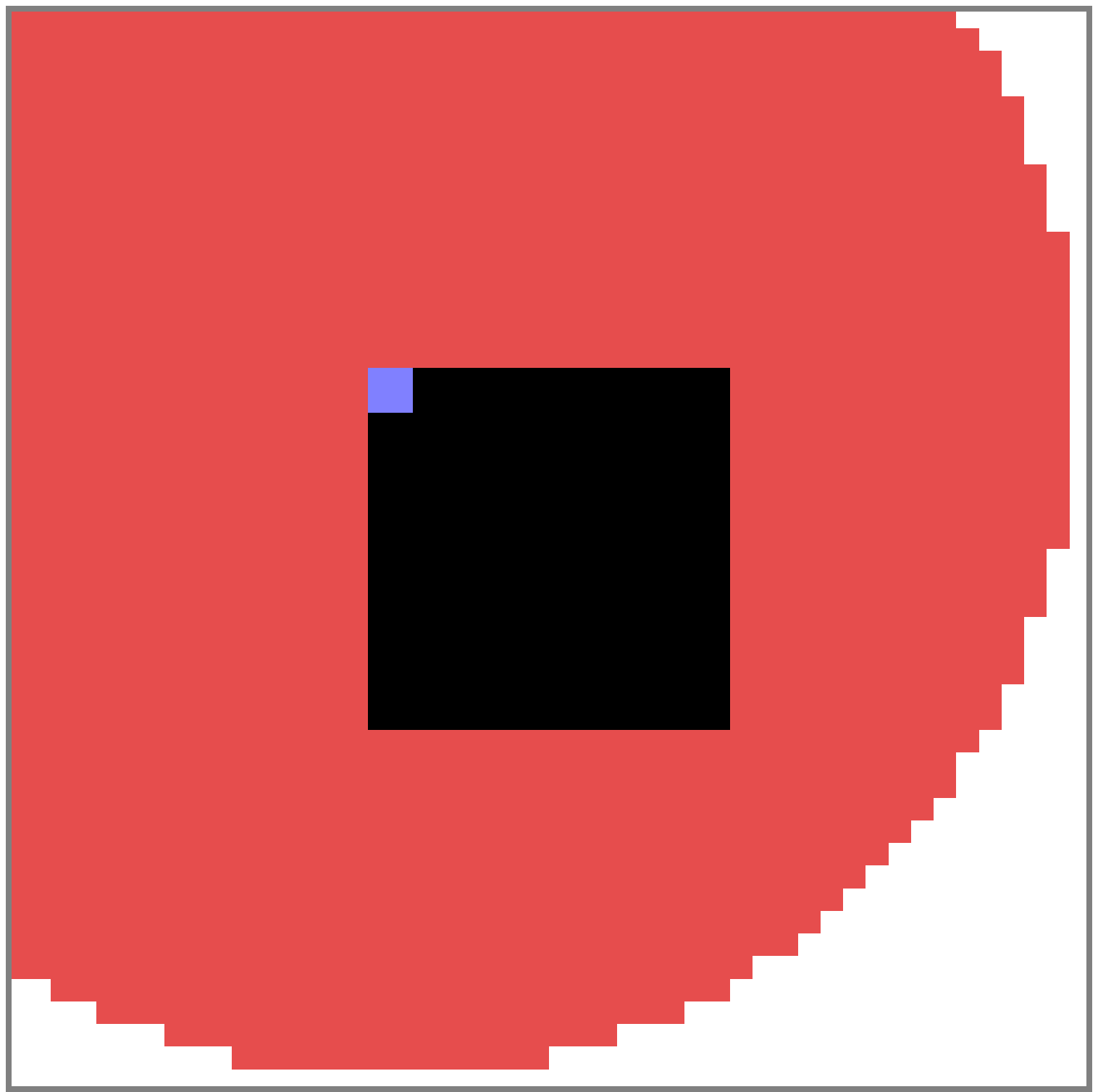}};
      \node (support_area_6) [below = 0.5cm of support_area_3] {\includegraphics[width=0.1\linewidth]{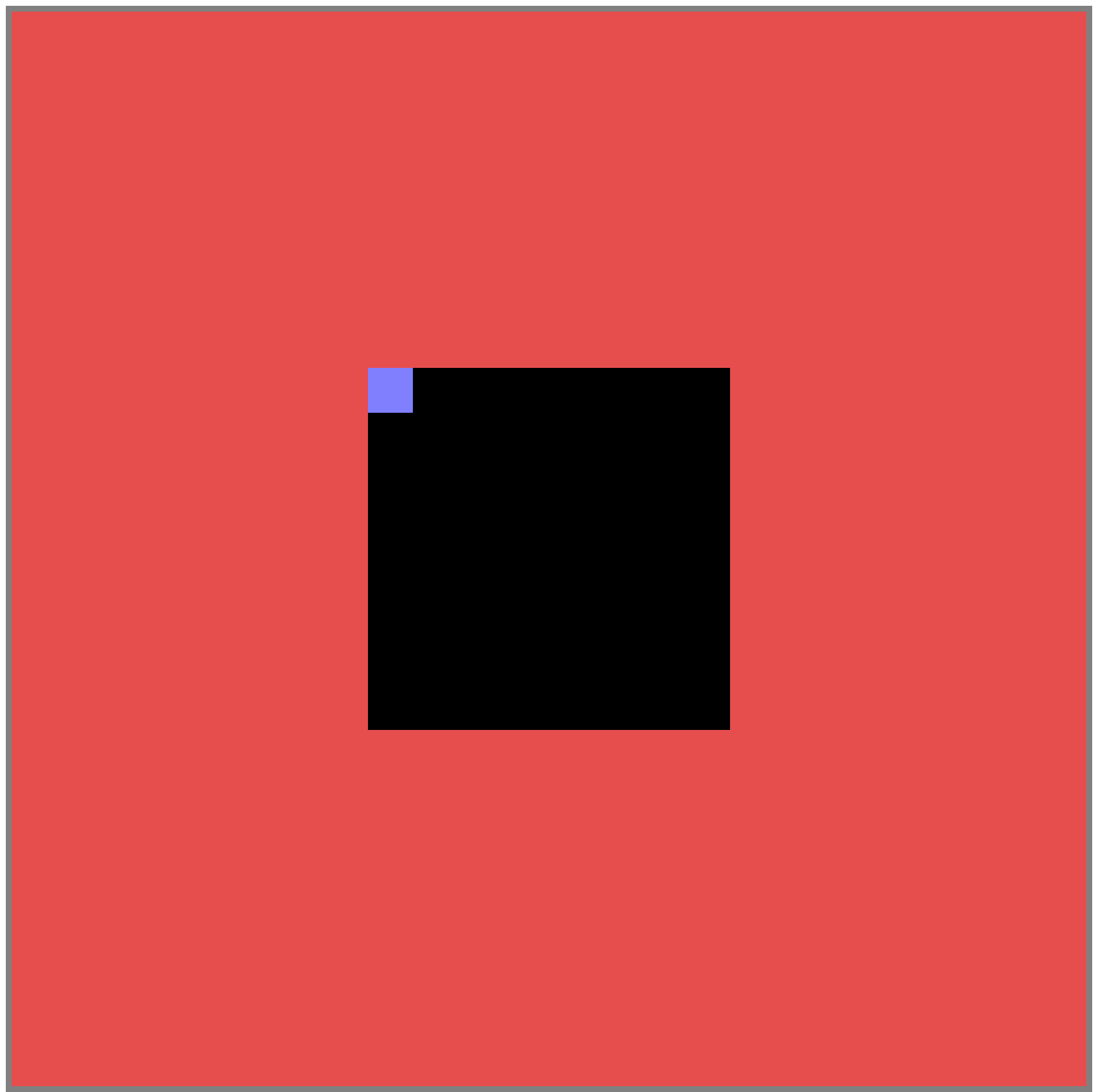}};
      
      \draw [arrow] (support_area_1) -- (support_area_2);
      \draw [arrow] (support_area_2) -- (support_area_3);
      \draw [arrow] (support_area_3) -- (support_area_4);
      \draw [arrow] (support_area_4) -- (support_area_5);
      \draw [arrow] (support_area_5) -- (support_area_6);
      
      \node (a) [below = 0.0cm of support_area] {(a)};
      \node (b) [right = 4.25cm of a] {(b)};
      
    \end{tikzpicture}  
  \end{center}        
    
    \caption{\small{(a) Expansion map of the support area. The support area is dynamically growing starting from pixels marked with the brightest red levels. Pixels marked with darker red levels are eventually incorporated later. (b) Example of support area expansion according to the expansion map. Pixels in black are missing, the currently processed patch is marked by blue colour.}}
    \label{fig:support_area}
\end{figure}

Furthermore, given the high spatial pixel correlation  within natural images, the prototypes $\bx_j$ that are more relevant for the SLP-E prediction lie more likely in the close vicinity of $\bx_0$. In fact, especially for periodic textures and regular structures, there may be no need to explore the entire support area $\S$. Therefore, we propose the support area to be dynamically growing from the location of the currently processing patch. Figure \ref{fig:support_area} shows an example of this dynamic expansion of the support area. The extension of the support area will be controlled by the selected profile.

This reconstruction layer, based on SLP-E with dynamically growing support area, will be referred to as intermediate dynamic layer (IDL) and the corresponding estimate is computed by inserting the reduced bandwidth in (\ref{bw_slpe}) into the estimator in (\ref{kmmse}), i.e.,
\begin{equation}
 \hat{\bx}_0^{\text{IDL}} = \sum_{j=1}^{M'} w_j^{\text{IDL}}(\by_0) \bx_j = \frac{1}{\nu}\sum_{j=1}^{M'} \exp\left(-\frac{1}{2} \frac{\|\by_j-\by_0\|^2}{\sigma^2 N_y} \right) \bx_j
 \label{slpe}
\end{equation}
where $\nu$ is the normalisation factor so that $\sum_{j=1}^{M'} w_j^{\text{IDL}} = 1$ and $M' \leq M$ denotes the number of templates $\by_j$ gathered from the expanded support area. The parameter $\sigma^2$ plays the role of a decay factor and it is set to 10 as in \cite{SLPE}. Simulations reveal that in stationary areas IDL yields reconstruction quality comparable to the full-featured K-MMSE (i.e. HQL) while requiring from around 5\% to 45\% of the computational time, depending on the extension of the support area.

\subsection{Basic reconstruction layer (BRL)}
\label{subsec:bl}
In visually flat regions, such as skies and walls, IDL can be further simplified with negligible effect on the reconstruction quality. 
Flat regions yield uniform weights (see (\ref{slpe})) so all patches $\bx_j$ are equally relevant for prediction. It follows that the dependency of the weights with the context $\by_0$ disappears and they can be set to a constant value. Moreover, in visually flat regions the available context $\by_0$ already carries sufficient information for performing the estimation. Thus, the support area $\S$ can be maximally reduced to contain $\by_0$ only. Furthermore, since flat regions lack structure, the spatial configuration of the prototypes is no longer relevant so vectors $\bx_j$ can be reduced to a set of scalars $x_j$. Given the reduction of $\S$ to $\by_0$, it follows that the scalar prototypes $x_j$ correspond to the pixels comprising $\by_0$. Thus, the corresponding estimate is computed as
\begin{equation}
 \hat{\bx}_0^{\text{BRL}} = \displaystyle{\sum_{j = 1}^{N_y} w^{\text{BRL}} x_j \bu =  \frac{1}{N_y}\sum_{j=1}^{N_y}\by_0(j) \bu}
 \label{mean}
\end{equation}
where $\bu = diag(I)$ with $I$ being an $N_x\times N_x$ identity matrix and $N_x$ denoting the dimensionality of the patch.
Note that the reconstruction in (\ref{mean}) is equivalent to filling the missing patch $\bx_0$ by the mean value of the pixels belonging to its context $\by_0$. In the following, this reconstruction method will be referred to as basic reconstruction layer (BRL) and it corresponds to context averaging. Given the simplicity of the estimator in (\ref{mean}), the computational time of BRL is slightly more than 1\% of the original K-MMSE computational time.

In this section, we have introduced the three reconstruction layers that comprise the proposed SK-MMSE estimator, namely:
\begin{enumerate}
 \item \textbf{HQL} corresponds to the full-featured K-MMSE approach.
 \item \textbf{IDL} consists of simplifying HQL by reducing the bandwidth matrix $\H$ to a scalar and suppressing its dependency with local image statistics.
 \item \textbf{BRL} consists of further simplifying IDL by reducing the support area to $\by_0$ and fixing the weights to a constant value.
\end{enumerate}

\section{Layer switching and profile selection}
\label{sec:profiles}

\begin{figure}[t]
    \centering
    \subfloat[\label{subfig-a:ca}]{%
      \includegraphics[width=0.95\linewidth]{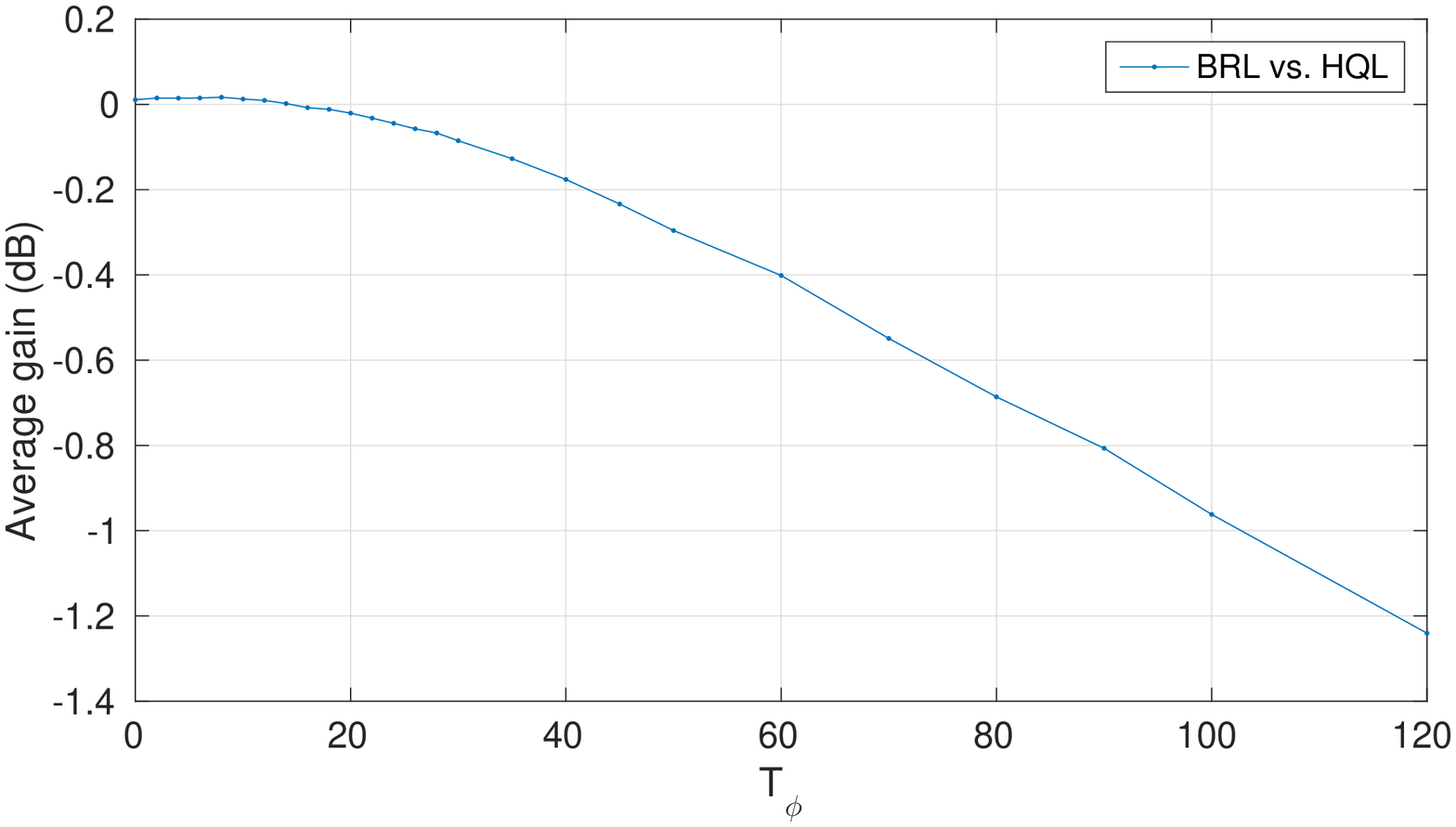}
    } \\    
    \subfloat[\label{subfig-b:ca}]{%
      \includegraphics[width=0.95\linewidth]{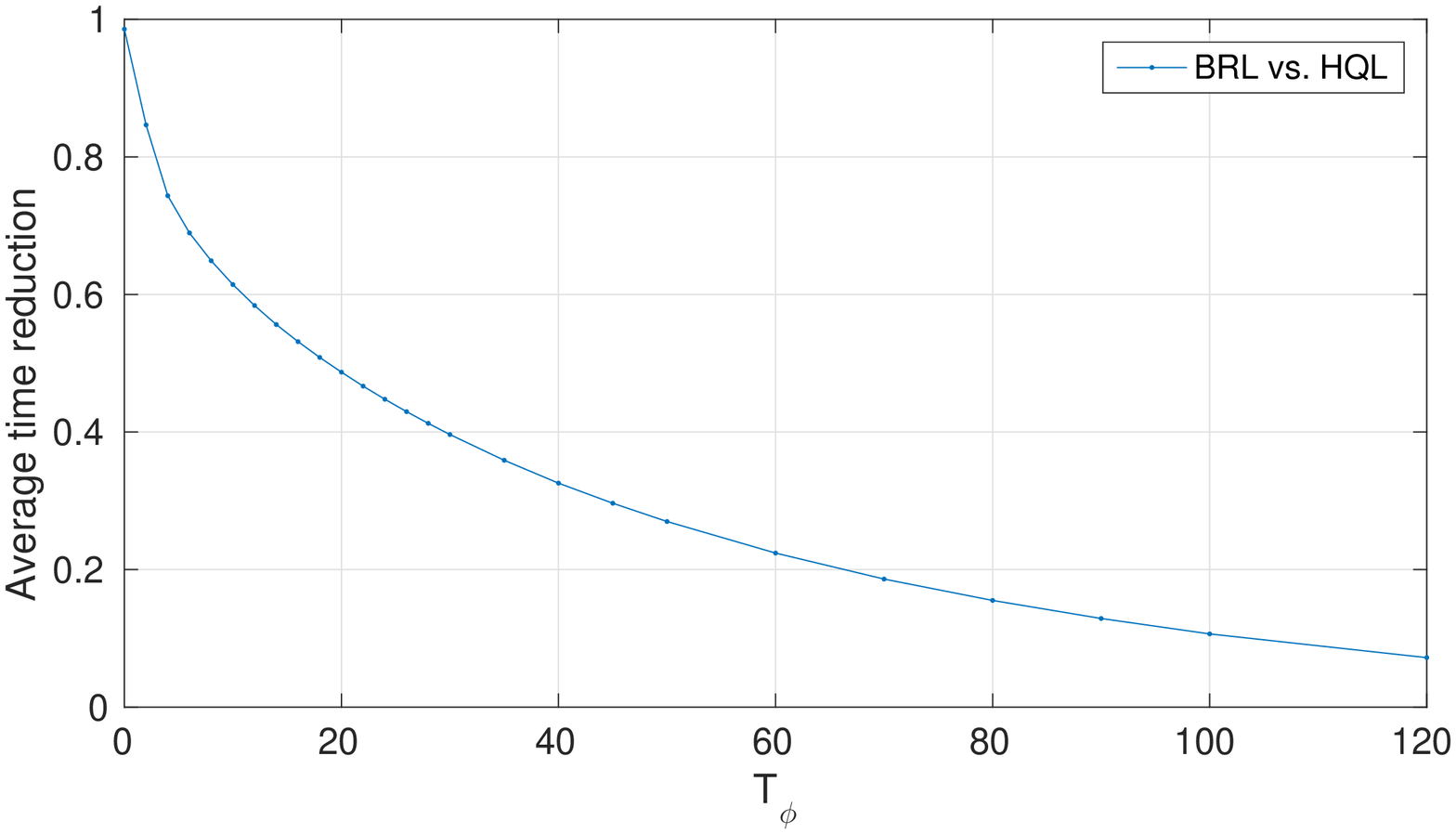}
    }
    \caption{\small{(a) Average PSNR gain (in dB) and (b) average time reduction with respect to K-MMSE for Tecnick image dataset as a function of the flatness threshold $T_{\phi}$.}}
    \label{fig:ca}
\end{figure}

\begin{figure}[t]
    \centering
    \subfloat[\label{subfig-a:slpe}]{%
      \includegraphics[width=0.95\linewidth]{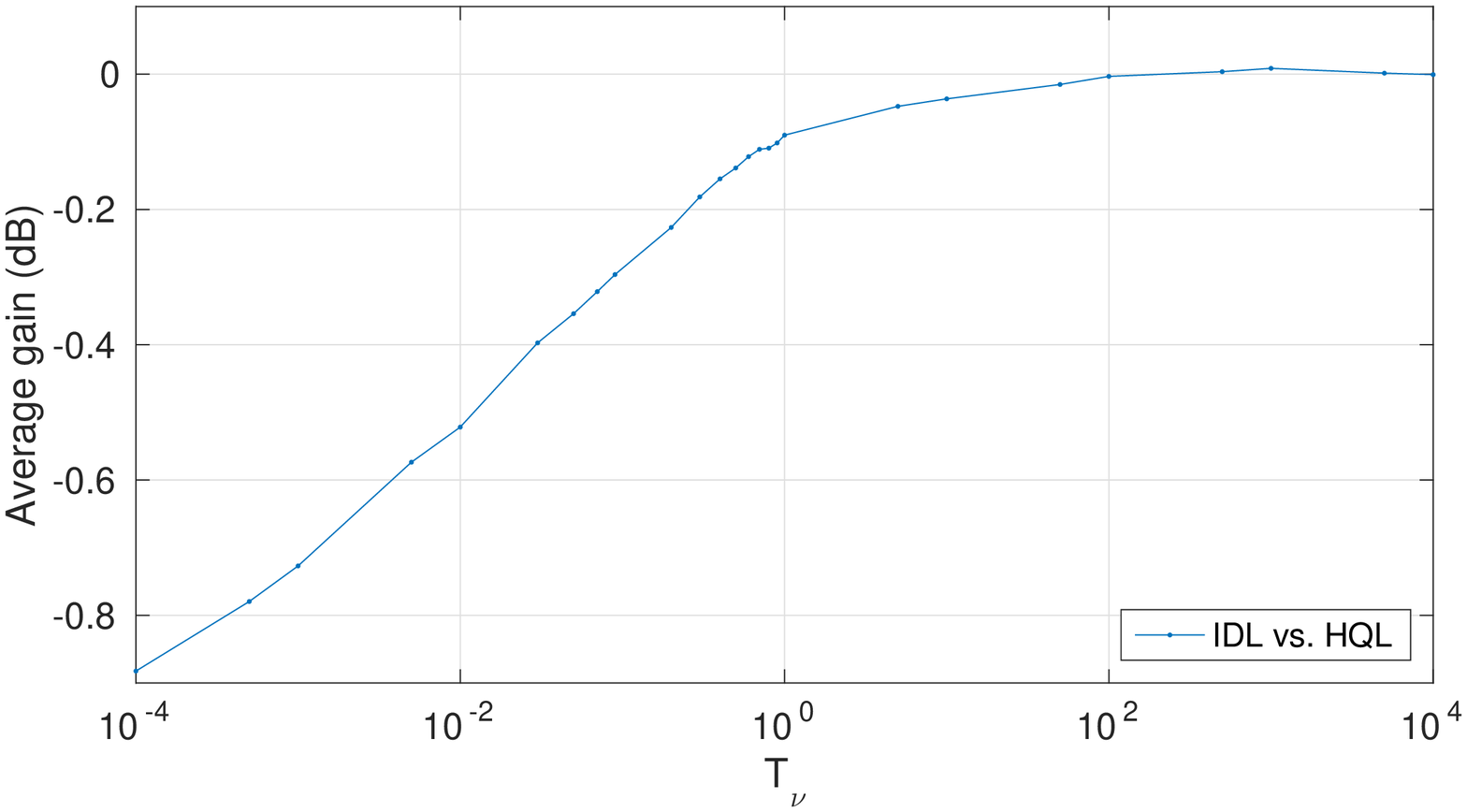}
    } \\    
    \subfloat[\label{subfig-b:slpe}]{%
      \includegraphics[width=0.95\linewidth]{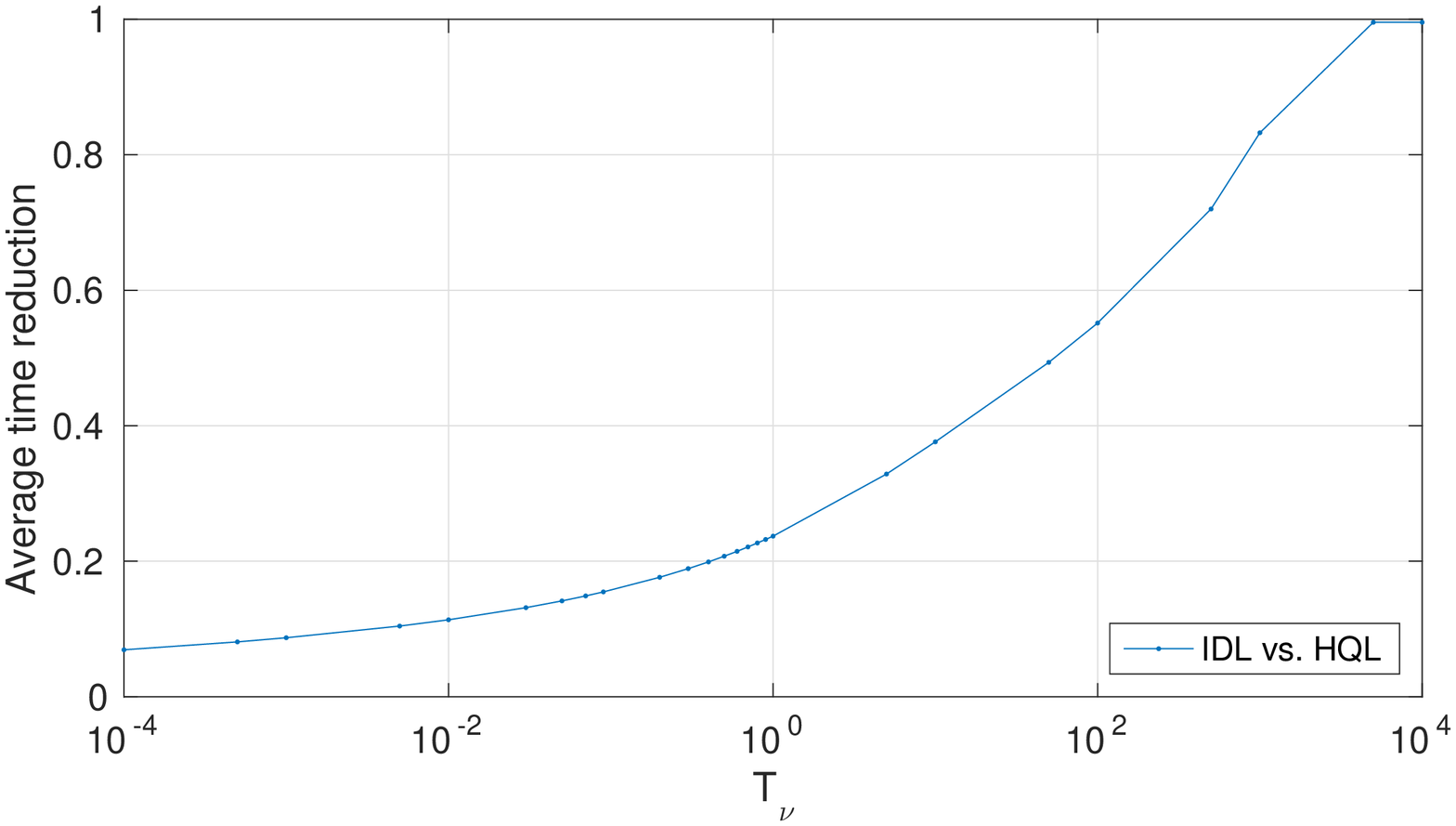}
    }
    \caption{\small{(a) Average PSNR gain (in dB) (b) average time reduction with respect to K-MMSE for Tecnick image dataset as a function of the normalisation threshold $T_{\nu}$. Logarithmic scale is used for more clarity.}}
    \label{fig:slpe}
\end{figure}

The goal of the proposed SK-MMSE is to accelerate the original \mbox{K-MMSE} reconstruction procedure with minimal effect on the resulting image quality. In order to do so, we propose a content adaptive layer selection scheme. This scheme is controlled by the so called SK-MMSE profiles that adjust the trade-off between the computational burden and the reconstruction quality. A profile therefore consists of a set of criteria that are applied to determine which layers are to be employed to perform the reconstruction. Given a patch to reconstruct, the algorithm evaluates the suitability of the different reconstruction layers and selects the best fitting one. This evaluation is sequential, going from the fast basic layer to the higher and more complex layers. Thus, if a layer is decided to be suitable, there is no need to test the remaining higher layers. The algorithm starts by testing the suitability of the fastest layer, i.e. BRL. Two assumptions are taken into consideration here:
\begin{enumerate}
 \item Given that BRL is the fastest layer, its suitability is always evaluated. Therefore, this evaluation should be simple and fast.
 \item As already mentioned, BRL is expected to perform well for visually flat areas. Thus, the visual flatness should be taken as the decision criterion. 
\end{enumerate}
In order to take into account both assumptions, we propose to use the dynamic range of $\by_0$ to measure the visual flatness $\phi$ of the available context $\by_0$, i.e.
\begin{equation}
 \phi(\by_0) = \max(\by_0) - \min(\by_0).
\end{equation}
The suitability evaluation of BRL therefore consists in comparing the current flatness to a certain threshold $T_{\phi}$. By adjusting the threshold, we can control the trade off between speed and reconstruction quality. Figure \ref{fig:ca}(a) shows the average PSNR gain (in dB) with respect to K-MMSE as a function of $T_{\phi}$.
We employ Tecnick image dataset \cite{Tecnick_v3}, designed for analysis and quality assessment, that is comprised by 100 images (600 $\times$ 600) and dispersed error pattern is applied. It follows that applying BRL for contexts with visual flatness below 20 yields virtually the same quality as full-featured K-MMSE reconstruction, i.e., HQL. For $\phi(\by_0)$ above 20 the performance starts to drop since the context can no longer be considered flat and context averaging is no longer a suitable reconstruction approach.

If the area under reconstruction is considered not flat, BRL leads inevitably to higher reconstruction error and, consequently, favours error propagation. In such cases, the suitability of the next layer (IDL) is to be evaluated. In order to do so, two assumptions are taken into account here:
\begin{enumerate}
 \item IDL exploits the high spatial correlation within natural images. In other words, it assumes that similar patches correspond to similar contexts. It follows that high quality reconstructions can be achieved if there are contexts $\by_j$ very close to $\by_0$.
 \item The higher the amount of contexts close to $\by_0$, the lower the reconstruction error. This relies on the fact that IDL consists of a weighted average of the available prototypes. A small number of prototypes close to $\bx_0$ can thus negatively affect the final reconstruction even though their corresponding (normalised) weights are large. 
\end{enumerate}
In order to take into account both facts, we propose to use the normalization factor $\nu$ (see (\ref{slpe})) to evaluate the suitability of IDL, i.e.,
\begin{equation}
 \nu(\by_0) = \sum_{j=1}^{M'} \exp \left( \frac{1}{2 \sigma^2} \frac{\| \by_j - \by_0 \|^2}{N_y} \right).
 \label{factor}
\end{equation}
Note that the higher the amount of good context candidates $\by_j$, the larger the normalization factor. The expression in (\ref{factor}) is equivalent to the sum of raw weights (without normalisation) and it indicates the total amount of data useful for prediction. Note that, unlike the normalised weights $w_j^{idl}$, a large raw weight reflects the high similarity between the spatial contexts $\by_j$ and $\by_0$. Thus, IDL will be evaluated as suitable if the corresponding normalization factor is larger than a given normalisation threshold $T_{\nu}$. It follows that IDL dynamically extends the support area $\S$ until the threshold criterion is met or until the support area is depleted, i.e., no more vectors $\bz_j = (\bx_j^t, \by_j^t)^t$ can be gathered from it. Figure \ref{fig:slpe}(a) shows the average PSNR gain (in dB) with respect to K-MMSE as a function of $T_{\nu}$ and Tecnick image dataset is employed again. It follows that the reconstruction quality starts to drop for $T_{\nu}$ below 100.

\subsection{SK-MMSE profiles}
\label{subsec:profiles}

In cases when both BRL (context averaging) and IDL (dynamic linear prediction) are evaluated as not suitable, i.e. $\phi(\by_0) > T_{\phi}$ and $\nu(\by_0) < T_{\nu}$, the missing patch is reconstructed by HQL (full-featured K-MMSE). The particular configuration of the two thresholds $T_{\phi}$ and $T_{\nu}$ will be referred to as profile. The trade-off between quality and speed can be controlled by setting the reconstruction profile. By studying the behaviour of the thresholds in Figs. \ref{fig:ca}(a) and \ref{fig:slpe}(a), it is observed that the reconstruction quality is approximately maintained intact for $T_{\phi} = 20$ and $T_{\nu} = 100$. Since this threshold configuration yields virtually no quality loss, it will be referred to as excellent profile in the following. Note that taking $T_{\phi} < 20$ and/or $T_{\nu} > 100$ will only slow down the SEC process and yield no reconstruction quality improvement. Taking the excellent profile as a reference, the computational complexity can be further reduced by relaxing the thresholds. Figures \ref{fig:ca}(b) and \ref{fig:slpe}(b) show the average time reduction of BRL and IDL with respect to HQL. It is shown that $T_{\nu}$ relaxation yields considerably better acceleration-distortion ratio that relaxing $T_{\phi}$. In other words, even though BRL reconstruction can be up to 30 times faster than IDL, the reconstruction quality tends to drop for visually heterogeneous areas. Therefore, in order to prevent error propagation, we will assume a fixed $T_{\phi}$ equal to 20 during the rest of the discussion. Thus, the trade-off between the complexity and the reconstruction quality can be controlled by adjusting $T_{\nu}$. In this paper, we propose three different SK-MMSE profiles, namely

\begin{figure}[t]    
    \includegraphics[width=1\linewidth]{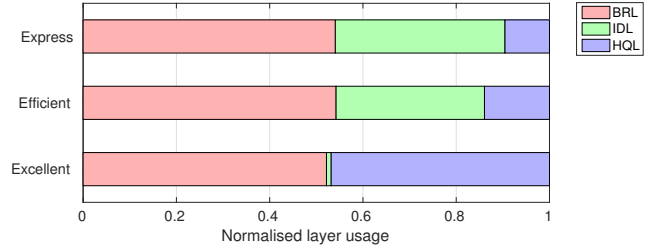}    
    \caption{\small{Average normalised layer usage for different reconstruction profiles. Kodak image dataset with dispersed error pattern is employed.}}
    \label{fig:profiles}
\end{figure}

\begin{enumerate}
 \item \textbf{Excellent} profile aims at obtaining the highest quality and, as mentioned, corresponds to $T_{\phi} = 20$ and $T_{\nu} = 100$.
 \item \textbf{Efficient} profile seeks for a balance between complexity and reconstruction quality and it corresponds to $T_{\phi} = 20$ and $T_{\nu} = 0.1$.
 \item \textbf{Express} profile focuses on maximal acceleration while keeping the quality decrease within moderate boundaries and it corresponds to $T_{\phi} = 20$ and $T_{\nu} = 0.01$.
\end{enumerate}
The settings for the three profiles are summarised in Table \ref{tab:profiles}. In addition, Fig. \ref{fig:profiles} shows the average usage of layers for the different profiles. In other words, it shows the distribution of the highest layer that is used during the EC process. It follows that BRL reconstruction is applied for slightly more than 50\% of the patches and it is approximately constant for all three profiles. This is the direct consequence of the fact that $T_{\phi}$ is set to the same value across the profiles. On the other hand, and as expected, the IDL usage increases as $T_{\nu}$ decreases. Finally, it is worth noticing that for the express profile less than 7\% of the patches (the most visually complex ones) are reconstructed by HQL.

\begin{table}[h]
	\centering
	\small
	\begin{tabular}{l|C|C|C}
		\hline\hline
		& \multicolumn{3}{c}{Profile}	\\
		& Express & Efficient & Excellent \\ \hline
		$T_{\phi}$ & 20 & 20 & 20\\
		$T_{\nu}$ & 0.01 & 0.1 & 100\\ \hline \hline
	\end{tabular}
	\caption{\small{Profile definitions.}}
	\label{tab:profiles}
\end{table}

\begin{figure*}[t]
	\centering	
	\includegraphics[width=1\linewidth,trim=4.5cm 9cm 5cm 4cm,clip]{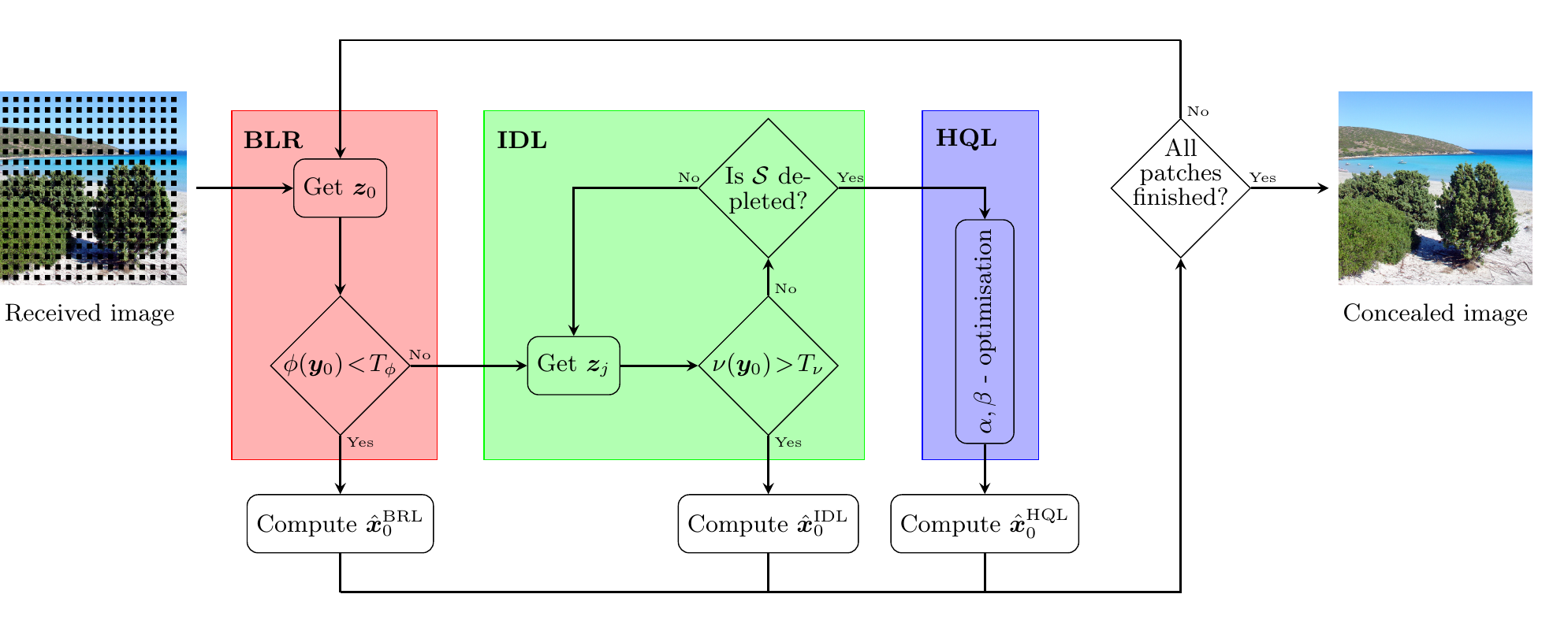}
	\caption{\small{Flow graph explaining SK-MMSE reconstruction algorithm.}}
	\label{fig:flowchart}
\end{figure*}


Note that, apart from the three proposed profiles, SK-MMSE offers a continuous range of reconstruction profiles. This flexibility of SK-MMSE can be further exploited by various broadcasting and streaming services for the automatic profile selection according to the characteristics of the employed transmission channels, available computational power or QoS demands.

\begin{table}[h]
	\centering
	\small
	\begin{tabular}{l | C | C | C}
		\hline\hline
		& \multicolumn{3}{c}{Reconstruction layer}	\\
		& BRL & IDL & HQL \\ \hline		
		$\frac{t_{\text{layer}}}{t_{\text{K-MMSE}}}$ & 1.30\% & up to 45.63\% & 100.01\% \\ \hline \hline
	\end{tabular}
	\caption{\small{Time ratio (in \%) with respect to K-MMSE for all three layers that comprise SK-MMSE.}}
	\label{tab:times}
\end{table}

The proposed algorithm is summarised by the flow graph in Fig. \ref{fig:flowchart}. It is shown that the SK-MMSE is a scalable EC procedure where lower layers feed information to higher layers. Related to the flow graph, Table \ref{tab:times} compares the computational burden, with respect to the full-featured \mbox{K-MMSE}, of all the stages that comprise SK-MMSE. It can be observed that BRL requires slightly more than 1\% of the computational time required by K-MMSE. On the other hand, since IDL iteratively expands the support area, its computational burden is not fixed. In case the entire support area needs to be exploited, IDL demands around 45\% of the computational time required by K-MMSE. Finally, HQL is slightly slower than K-MMSE due to the fact that profile checks to thresholds $T_{\phi}$ and $T_{\nu}$, that are not considered in the original K-MMSE framework, need to be evaluated.

\section{Simulation results}
\label{sec:results}

In order to thoroughly evaluate the performance of the proposed algorithm, we have conducted series of tests on both still images and video sequences. The obtained results are presented and discussed in the next subsections.

\subsection{Experiments with still images}
\label{subsec:images}

In order to better take into account the perceptual quality, the peak signal-to-noise ratio (PSNR) as well as the structural similarity (SSIM) index \cite{SSIM} are used to measure the performance. Unlike PSNR, which objectively measures the squared reconstruction error, SSIM additionally take into consideration the differences in contrast and structure. Block losses with dispersed and random pattern, both corresponding to approximately 25\%, have been simulated. Moreover, since the threshold behaviour analysis has been carried out employing Tecnick database, we will test the performance of the proposed SK-MMSE technique on Kodak image dataset \cite{Kodak} comprised by 24 images (768$\times$512) with bit-depth of 8 bits.

We have compared our technique with other state-of-the-art EC algorithms, namely frequency selective extrapolation (FSE) \cite{FOFSE}, edge recovery based on visual clearness (EVC) \cite{EVC}, content adaptive EC (CAD) \cite{CAD}, sparse linear prediction with exponential weights (SLP-E) \cite{SLPE}, multi-directional interpolation (MDI) \cite{MDI}, and bilateral filtering (BLF) \cite{BLF}.

\begin{figure}[t]
    \centering
    \subfloat[\label{subfig-a:dispersed_psnr}]{%
      \includegraphics[width=0.95\linewidth]{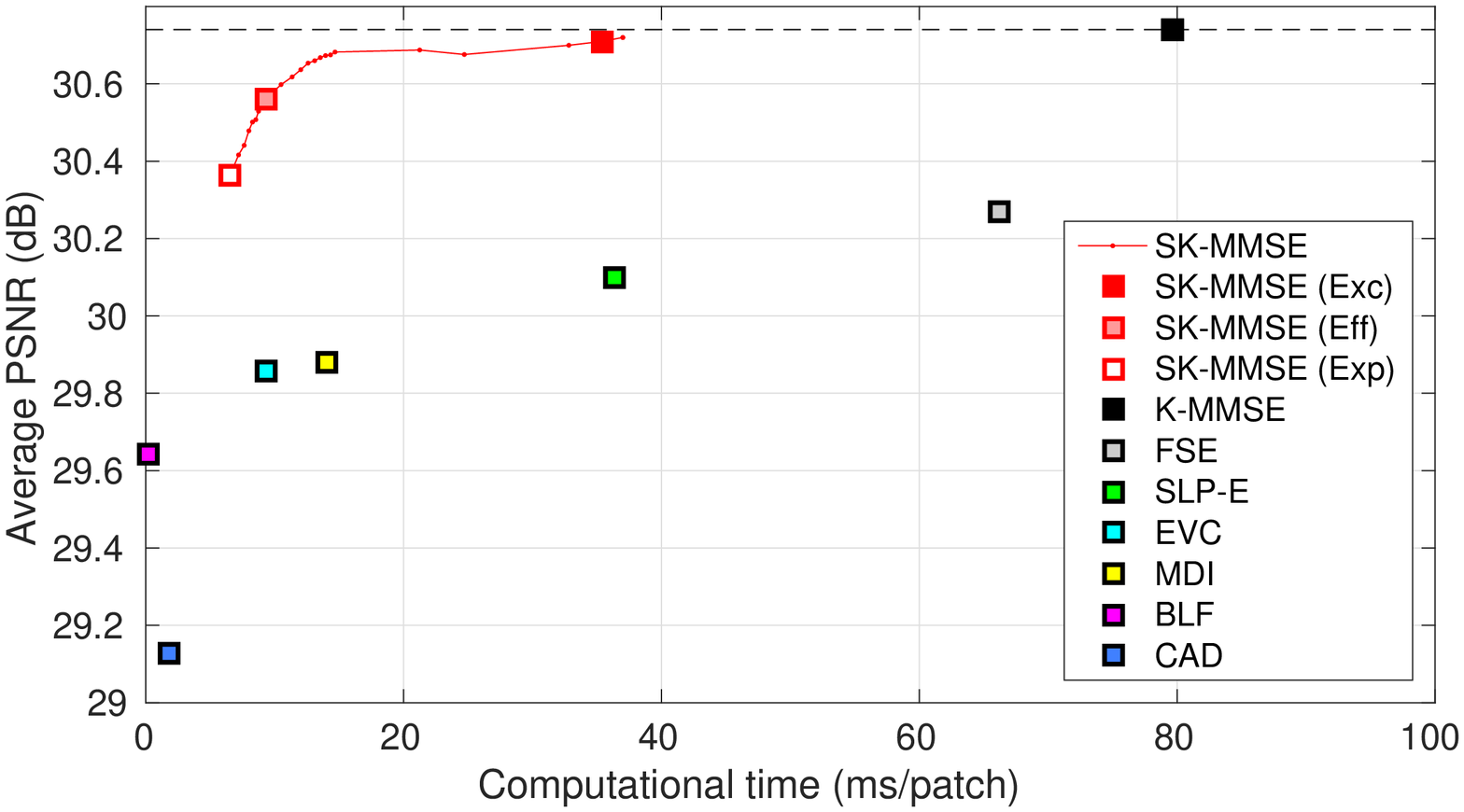}
    } \\    
    \subfloat[\label{subfig-b:dispersed_ssim}]{%
      \includegraphics[width=0.95\linewidth]{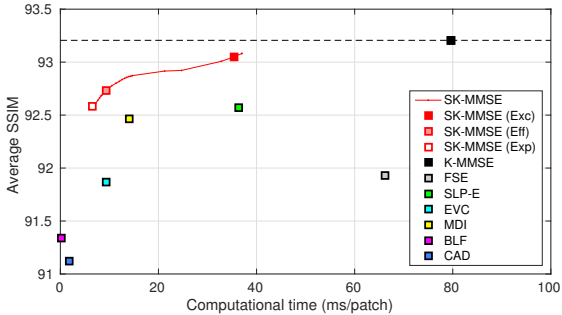}
    }
    \caption{\small{(a) Average PSNR (in dB)  and (b) average SSIM ($\times$ 100) as a function of processing time (in milliseconds per patch) for different SEC techniques. Kodak image dataset with \textit{dispersed} error pattern is employed}}
    \label{fig:dispersed_results}
\end{figure}

Figure \ref{fig:dispersed_results}(a) shows the average PSNR (in dB) as a function of the processing time for different algorithms and employing the dispersed error pattern. The results for the three \mbox{SK-MMSE} profiles are also indicated. The performance curve for \mbox{SK-MMSE} is obtained by continuously adjusting the threshold $T_{\nu}$ according to the mechanism described previously. It is observed that SK-MMSE can maintain the high reconstruction quality of K-MMSE while reducing the processing time more than twice. A six times reduction yields a quality decrease of less than 0.1dB with respect to K-MMSE. In addition, SK-MMSE is able to reduce the computational burden more than 10 times and it still outperforms other state-of-the-art techniques. This performance behaviour is also corroborated by SSIM measurements, as shown in Fig. \ref{fig:dispersed_results}(b).

\begin{figure}[t]
    \centering
    \subfloat[\label{subfig-a:random_psnr}]{%
      \includegraphics[width=0.95\linewidth]{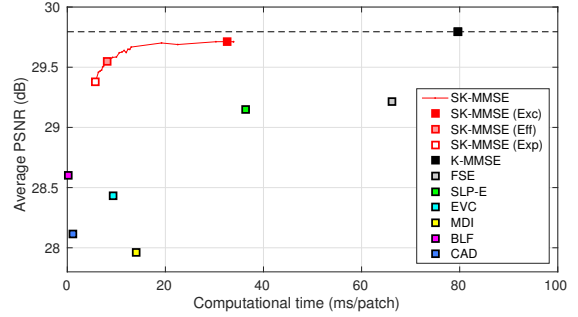}
    } \\    
    \subfloat[\label{subfig-b:random_ssim}]{%
      \includegraphics[width=0.95\linewidth]{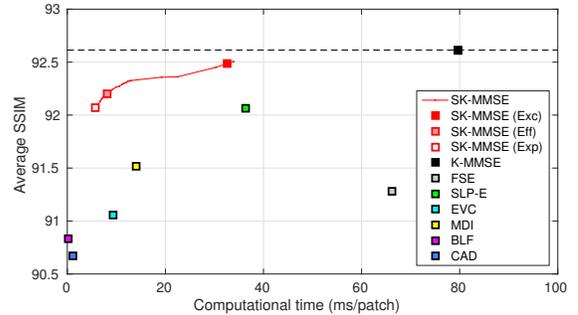}
    }
    \caption{\small{(a) Average PSNR (in dB)  and (b) average SSIM ($\times$ 100) as a function of processing time (in milliseconds per patch) for different SEC techniques. Kodak image dataset with \textit{random} error pattern is employed}}
    \label{fig:random_results}
\end{figure}

The results for the random error pattern are shown in Fig. \ref{fig:random_results}. The average PSNR (in dB) as a function of the processing time for different algorithms is shown in Fig. \ref{fig:random_results}(a). Again, it is shown that negligible losses in reconstruction quality are observed while the computational burden is reduced more than twice. Furthermore, a 10 times reduction yields only a moderate PSNR decrease and the resulting reconstruction quality still outperforms other state-of-the-art techniques by around 0.4dB. The SSIM measurements, shown in Fig. \ref{fig:random_results}(b), also corroborate this performance behaviour.

\begin{table}[b]
	\centering
	\small
	\begin{tabular}{l|cc|cc}
		\hline\hline
		& \multicolumn{2}{c}{Time reduction ($\times$)} & \multicolumn{2}{c}{Average gain (dB)} \\
		Profile & Dispersed & Random & Dispersed & Random \\ \hline
		Express   & 12.19 & 13.84 & -0.38 & -0.41 \\
		Efficient & 8.57 & 9.82 & -0.18 & -0.24 \\
		Excellent & 2.25 & 2.24 & -0.03 & -0.08 \\ \hline \hline
	\end{tabular}
	\caption{\small{Performance of the different SK-MMSE reconstruction profiles. Average computational time reductions and average PSNR gains, both with respect to the full-featured K-MMSE, are indicated for dispersed and random error pattern. Kodak image dataset is employed.}}
	\label{tab:results}
\end{table}

For the three reconstruction profiles, the average PSNR gain and the corresponding computational time reductions with respect to K-MMSE are shown in Table \ref{tab:results}. Both dispersed and random error patterns are considered. It is observed that the excellent reconstruction profile yields virtually no quality losses while reducing the computational time to approximately 44\%. The efficient reconstruction profile exhibits a slight quality loss of around 0.2dB while accelerating the reconstruction process up to 10 times. The computational time can be further reduced down to 7\% when using the express profile at the expense of a moderate quality loss of around 0.4dB.

\begin{figure*}[t]
  \begin{center}
      \begin{tikzpicture}
	\node (a) {\includegraphics[width=0.24\linewidth]{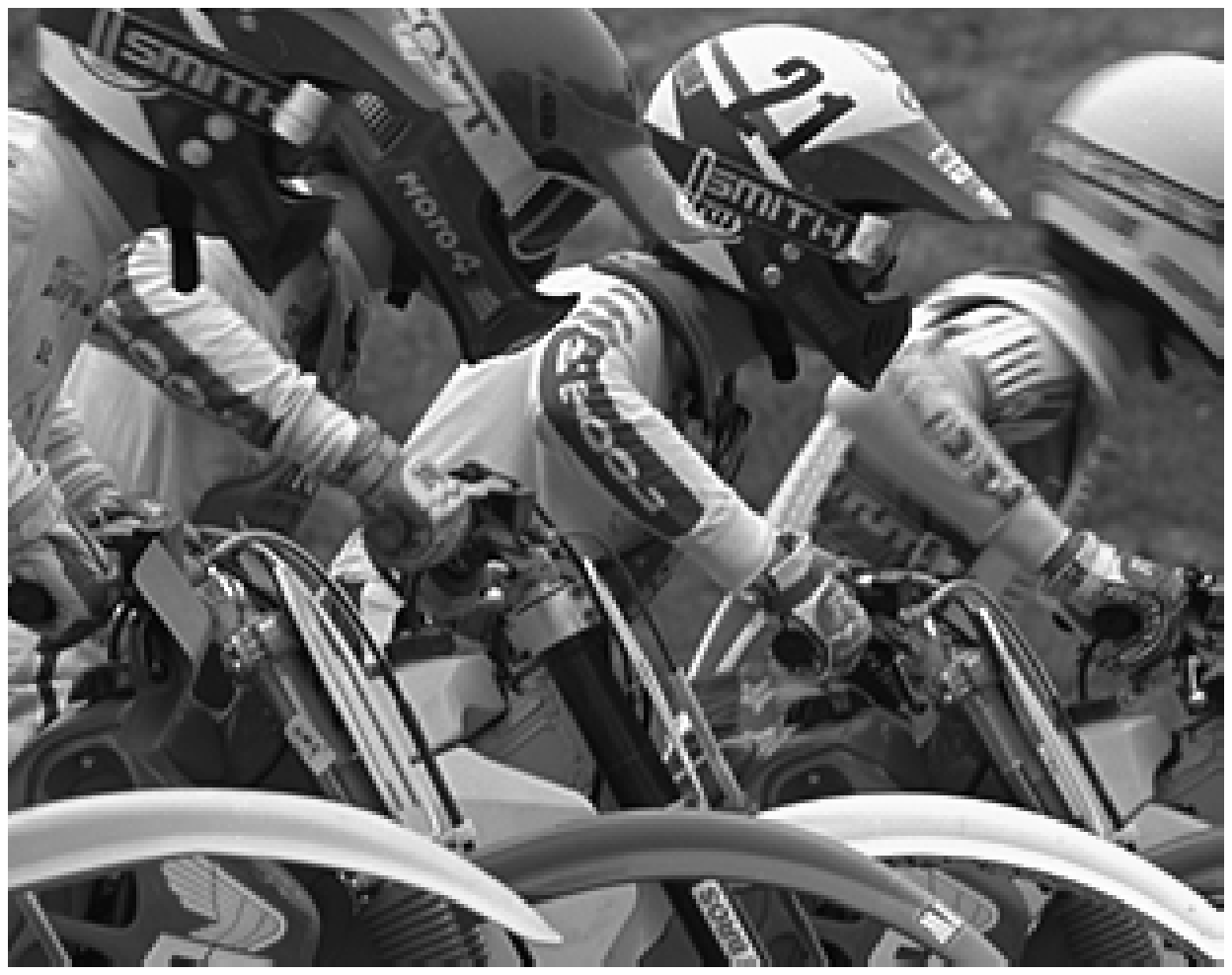}};
	\node (b) [right = -0.15cm of a] {\includegraphics[width=0.24\linewidth]{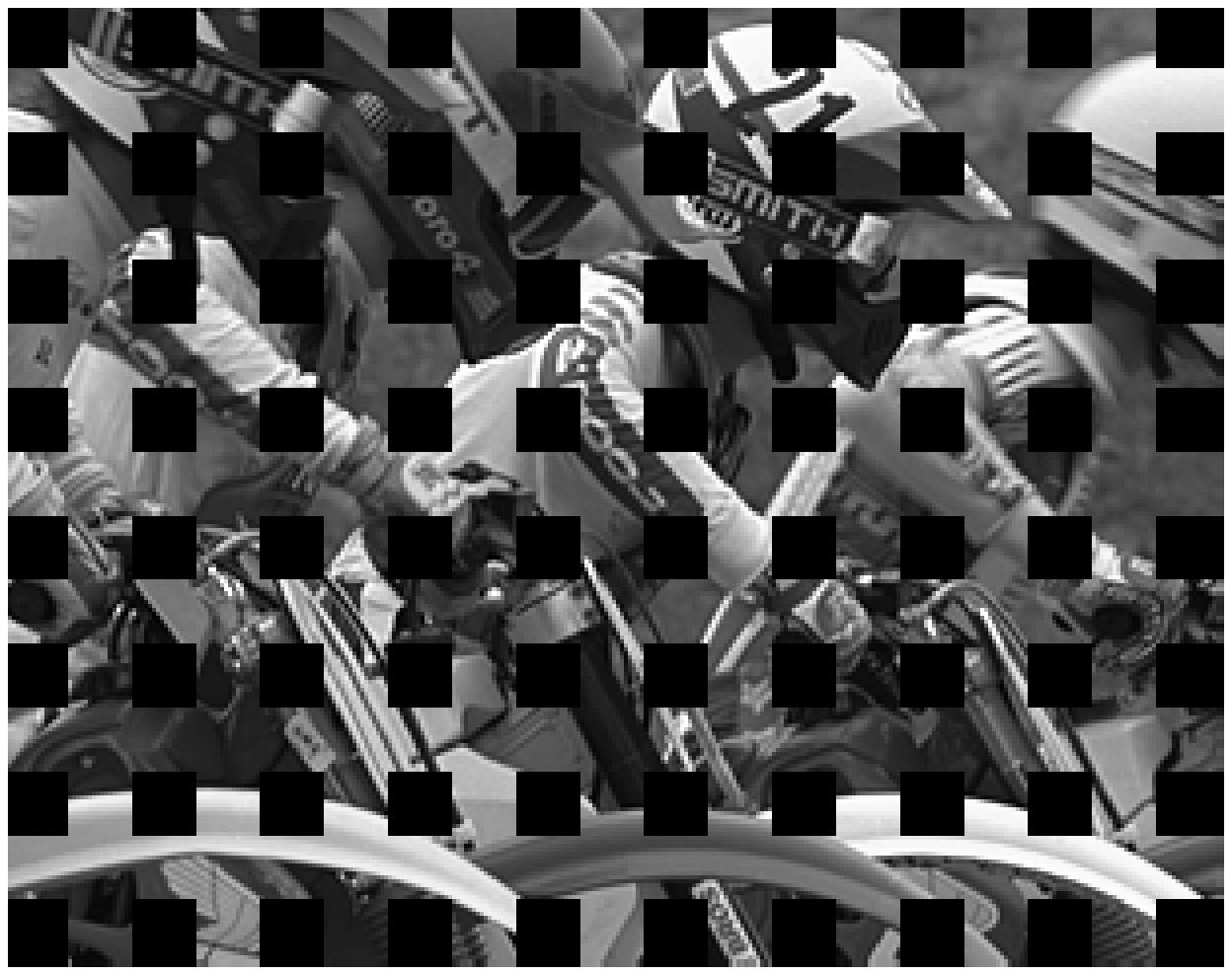}};
	\node (c) [right = -0.15cm of b] {\includegraphics[width=0.24\linewidth]{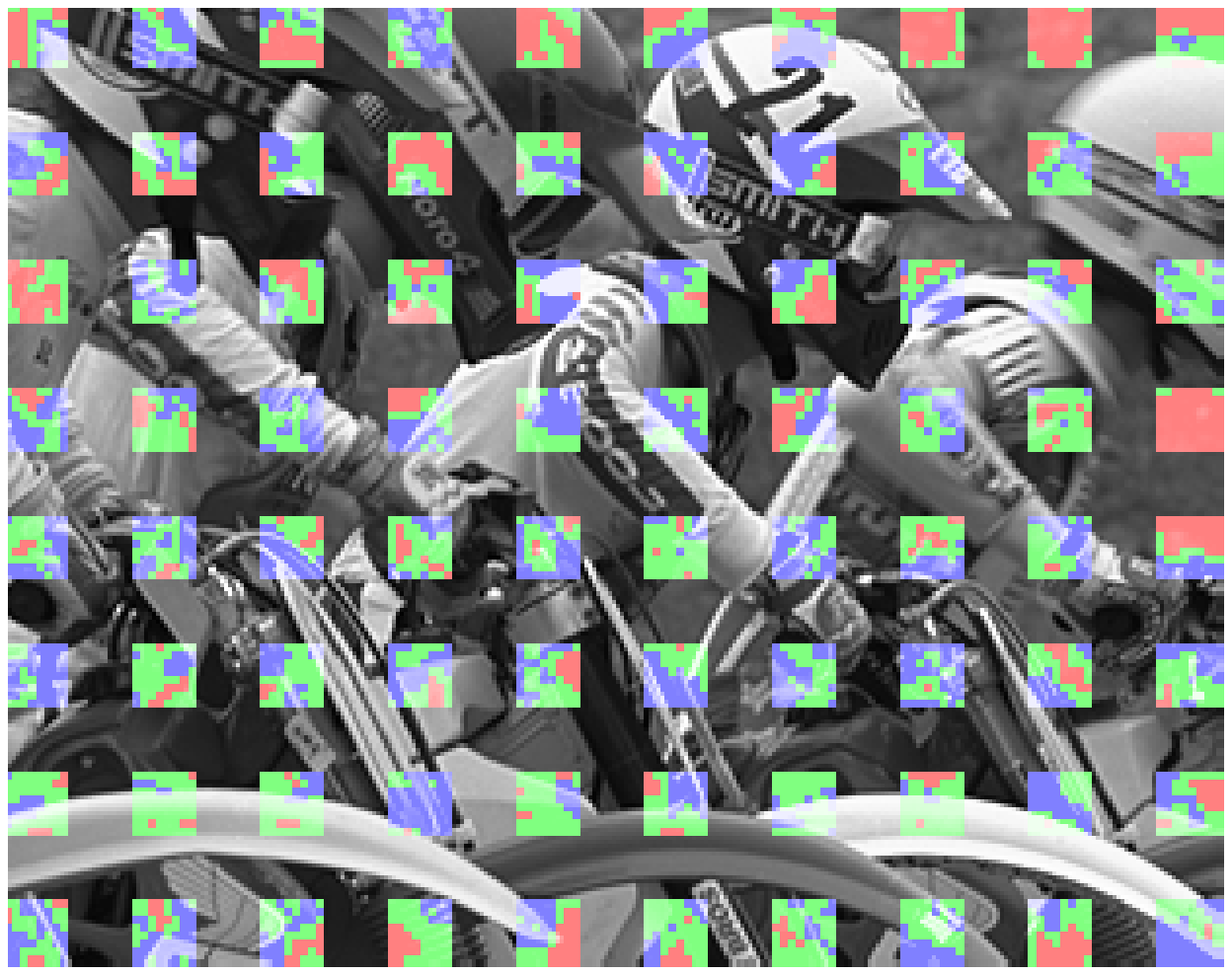}};
	\node (d) [right = -0.15cm of c] {\includegraphics[width=0.24\linewidth]{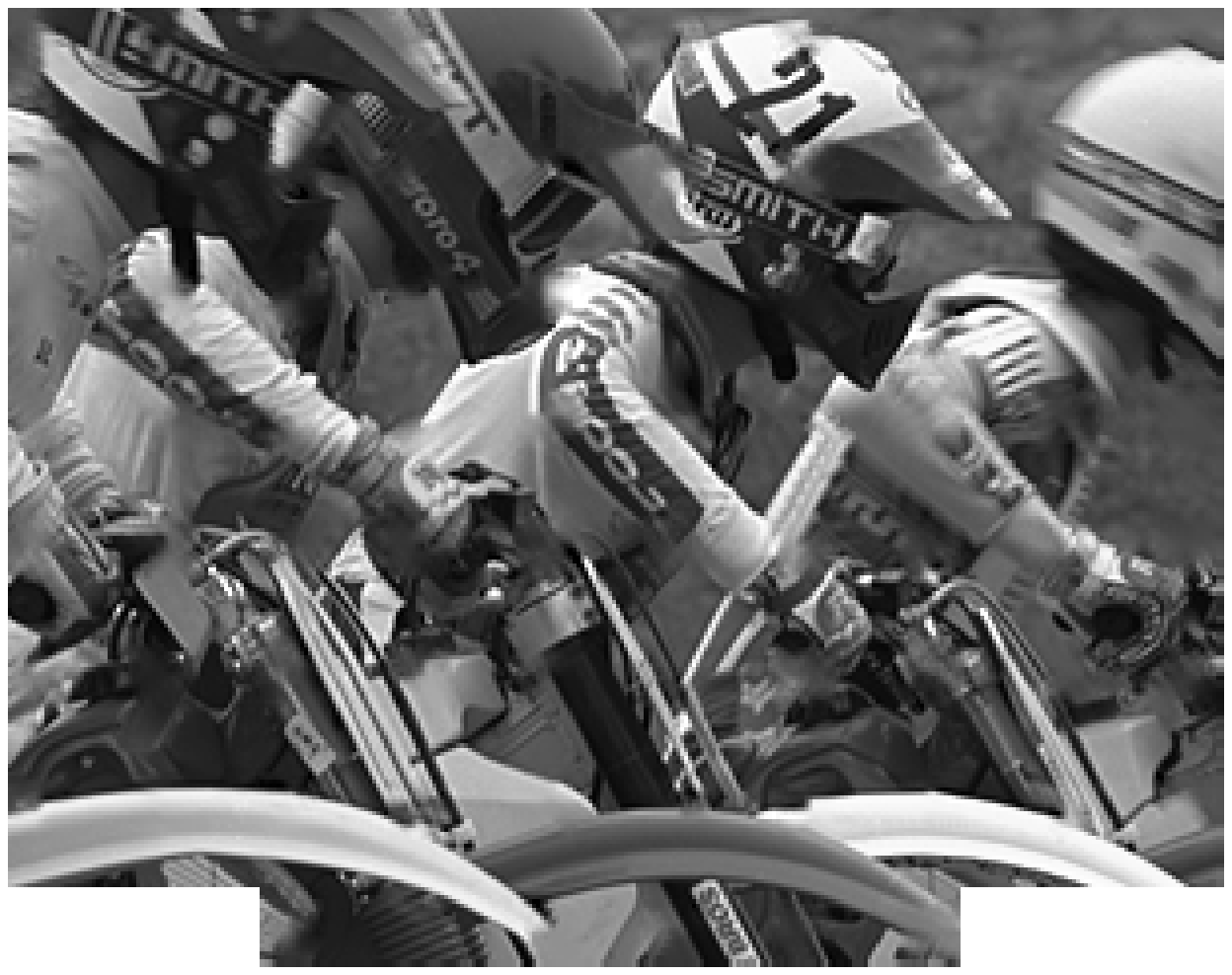}};
	
	\node (a_label) [below = -0.1cm of a] {\small{(a)}};
	\node (b_label) [below = -0.1cm of b] {\small{(b)}};
	\node (c_label) [below = -0.1cm of c] {\small{(c)}};
	\node (d_label) [below = -0.1cm of d] {\small{(d)}};
	
	\node (e) [below = 0.25cm of a] {\includegraphics[width=0.24\linewidth]{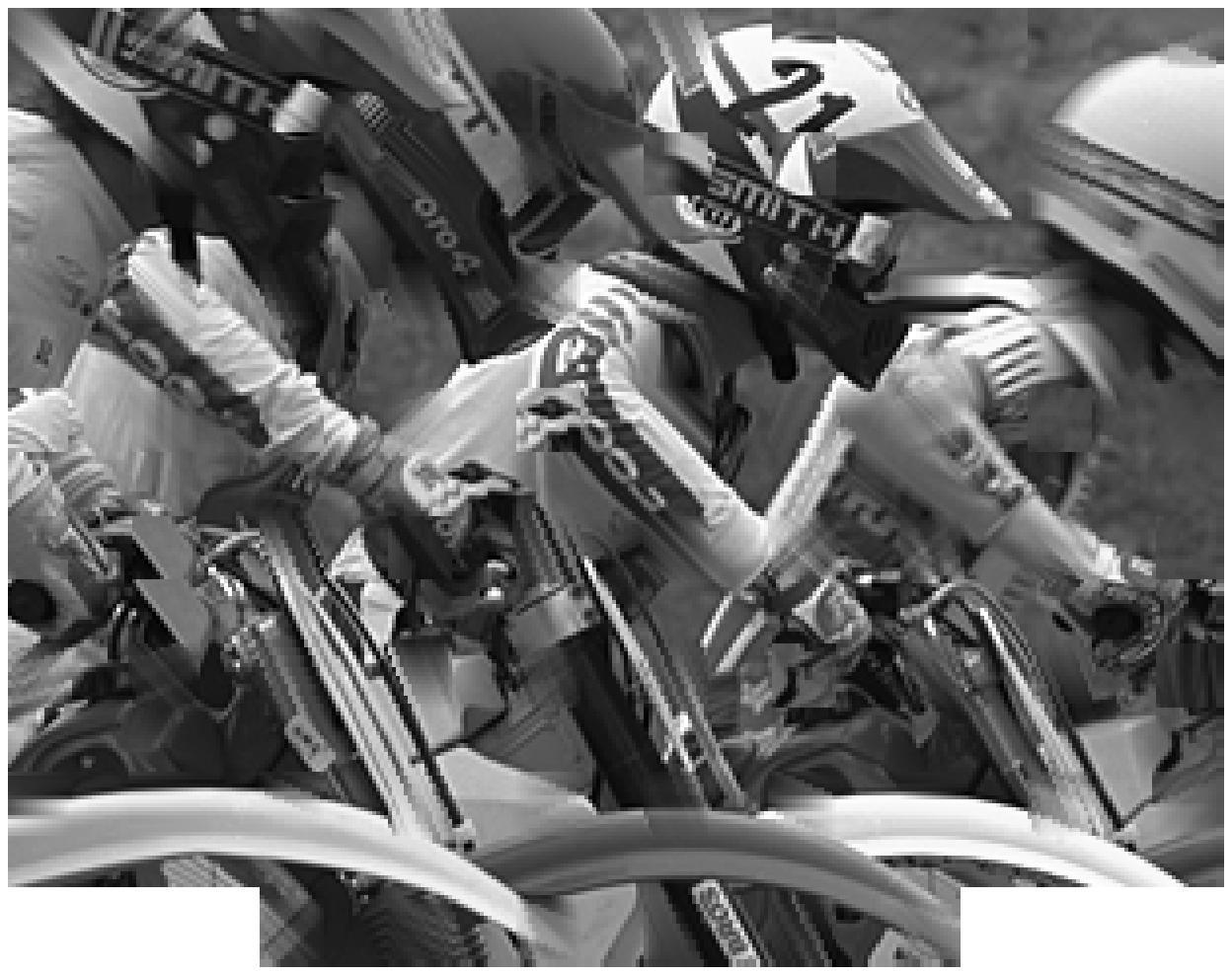}};
	\node (f) [right = -0.15cm of e] {\includegraphics[width=0.24\linewidth]{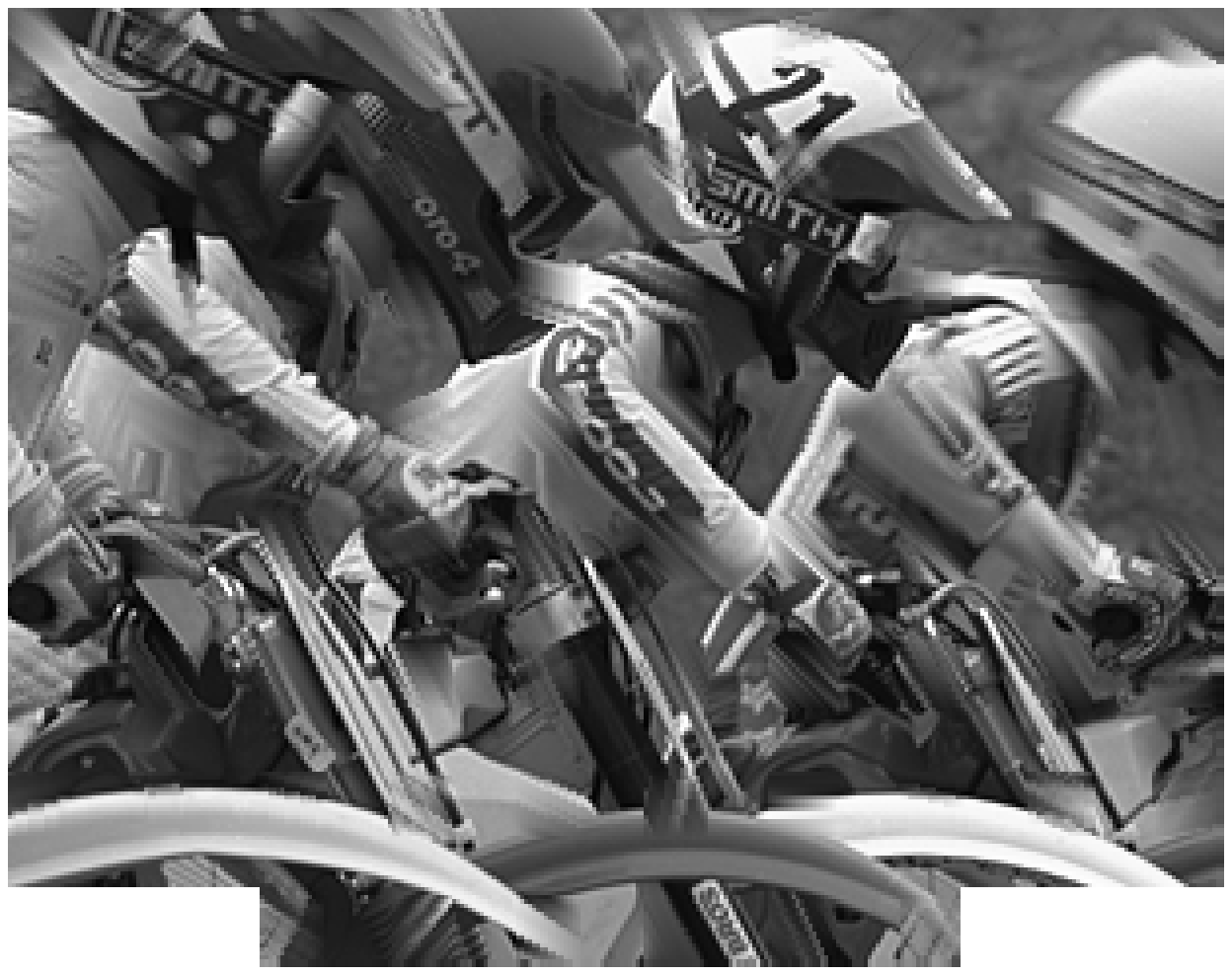}};
	\node (g) [right = -0.15cm of f] {\includegraphics[width=0.24\linewidth]{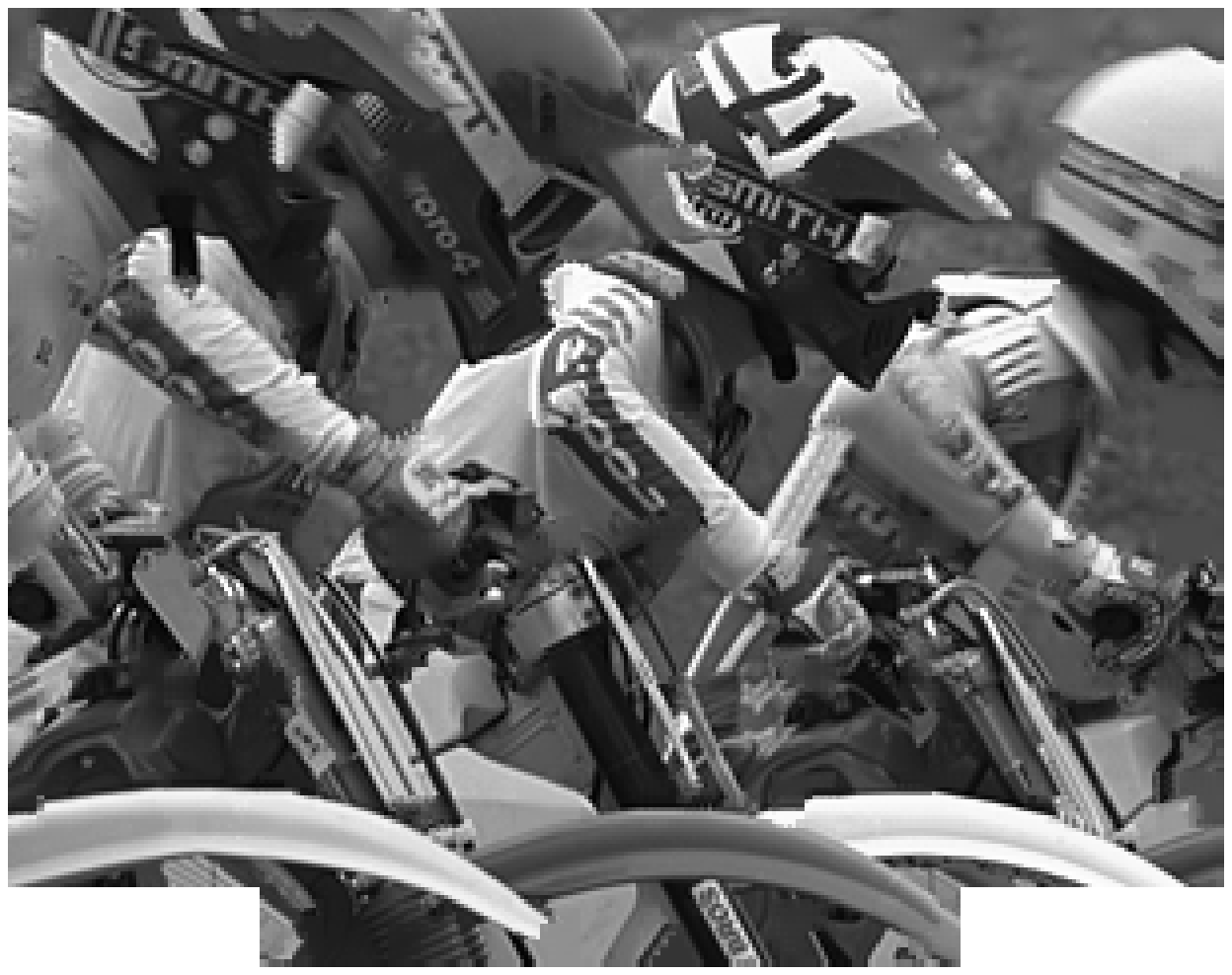}};
	\node (h) [right = -0.15cm of g] {\includegraphics[width=0.24\linewidth]{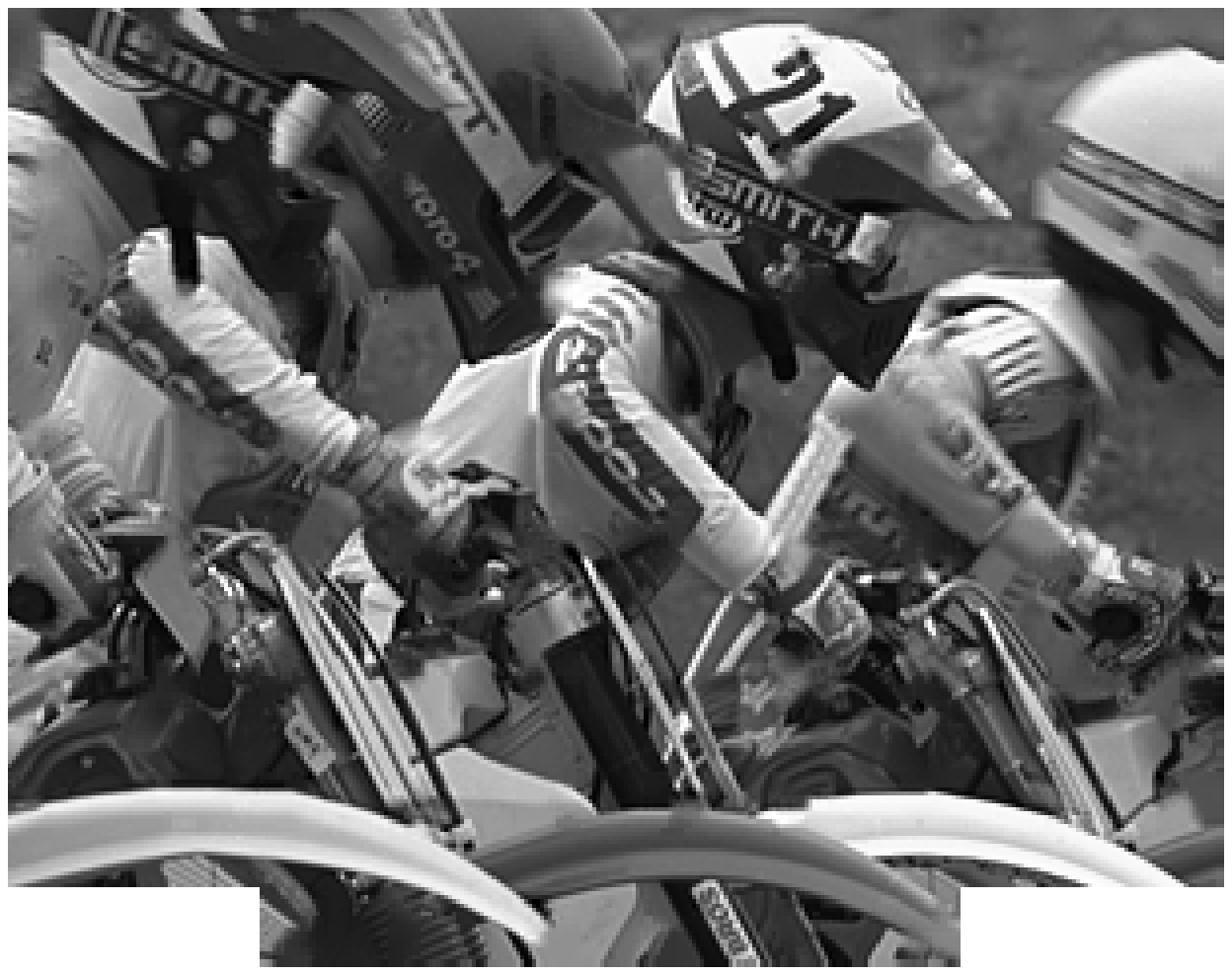}};
		
	\node (e_label) [below = -0.1cm of e] {\small{(e)}};
	\node (f_label) [below = -0.1cm of f] {\small{(f)}};
	\node (g_label) [below = -0.1cm of g] {\small{(g)}};
	\node (h_label) [below = -0.1cm of h] {\small{(h)}};
	
	\node (d_psnr) [below left = -0.45cm and -1.15cm of d] {\scriptsize{\textbf{26.33dB}}};
	\node (e_psnr) [below left = -0.45cm and -1.15cm of e] {\scriptsize{\textbf{24.51dB}}};
	\node (f_psnr) [below left = -0.45cm and -1.15cm of f] {\scriptsize{\textbf{25.09dB}}};
	\node (g_psnr) [below left = -0.45cm and -1.15cm of g] {\scriptsize{\textbf{25.26dB}}};
	\node (h_psnr) [below left = -0.45cm and -1.15cm of h] {\scriptsize{\textbf{26.23dB}}};
	
	\node (d_time) [below right = -0.45cm and -1.15cm of d] {\scriptsize{\textbf{79.65ms}}};
	\node (e_time) [below right = -0.45cm and -1.15cm of e] {\scriptsize{\textbf{1.26ms}}};
	\node (f_time) [below right = -0.45cm and -1.15cm of f] {\scriptsize{\textbf{9.29ms}}};
	\node (g_time) [below right = -0.45cm and -1.15cm of g] {\scriptsize{\textbf{36.34ms}}};
	\node (h_time) [below right = -0.45cm and -1.15cm of h] {\scriptsize{\textbf{21.34ms}}};
	
      \end{tikzpicture}
    \end{center}
    \vspace{-0.5cm}
    \caption{\small{Comparison of reconstructions obtained by different techniques and employing \textit{dispersed} error pattern. (a) Original image from Kodak database, frame \#5. (b) Received frame. (c) Layer distribution for SK-MMSE efficient profile. Patches reconstructed by BRL are in red, by IDL in green and by HQL in blue. Reconstruction by (d) K-MMSE, (e) CAD, (f) EVC, (g) SLP-E, (h) SK-MMSE (efficient profile). The corresponding PSNR values (in dB) and computational times (in milliseconds per patch) are also indicated. Best to be viewed enlarged on a screen.}}
    \label{fig:example_dispersed}  
\end{figure*}

\begin{figure*}[h!]
  \begin{center}
      \begin{tikzpicture}
	\node (a) {\includegraphics[width=0.24\linewidth]{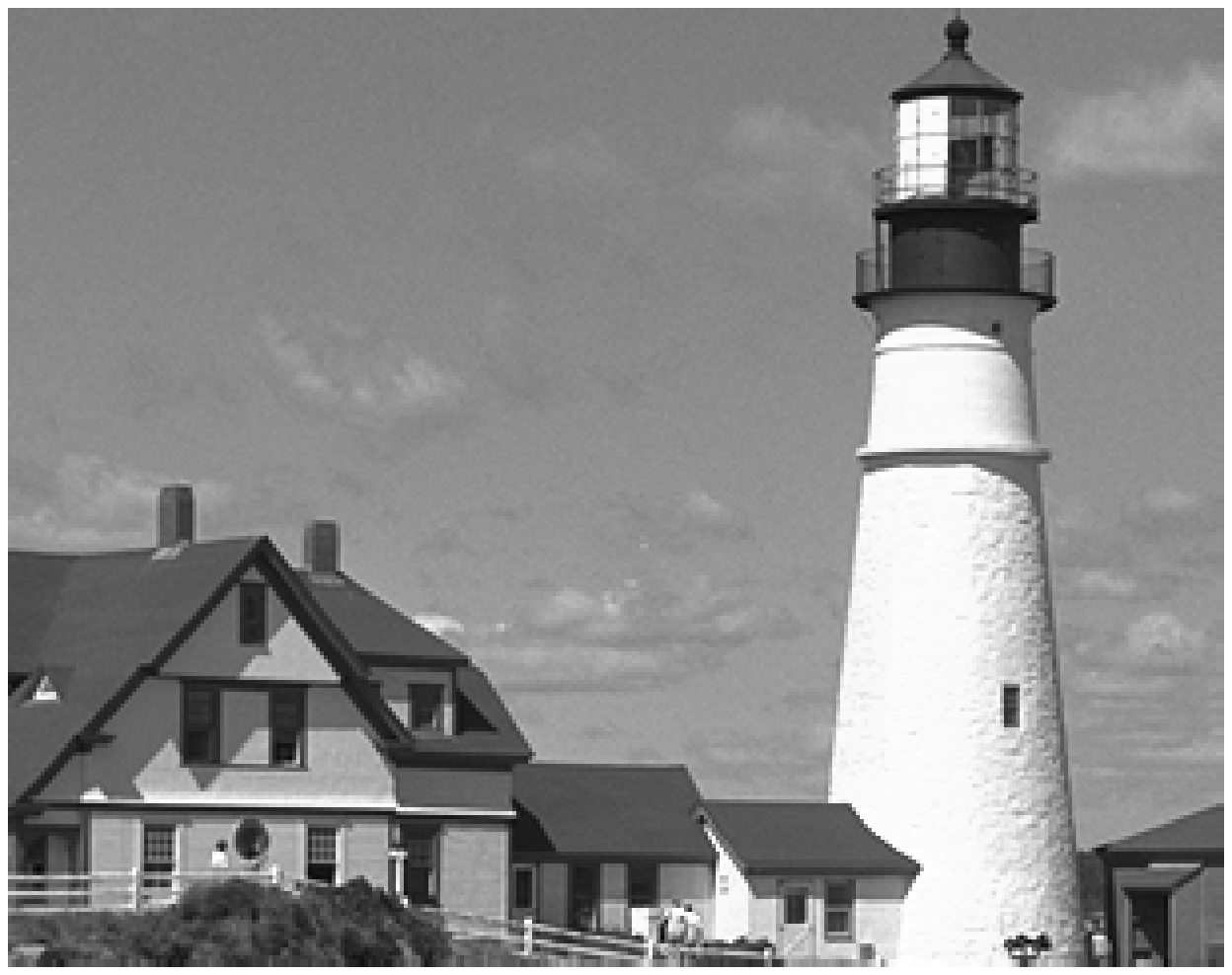}};
	\node (b) [right = -0.15cm of a] {\includegraphics[width=0.24\linewidth]{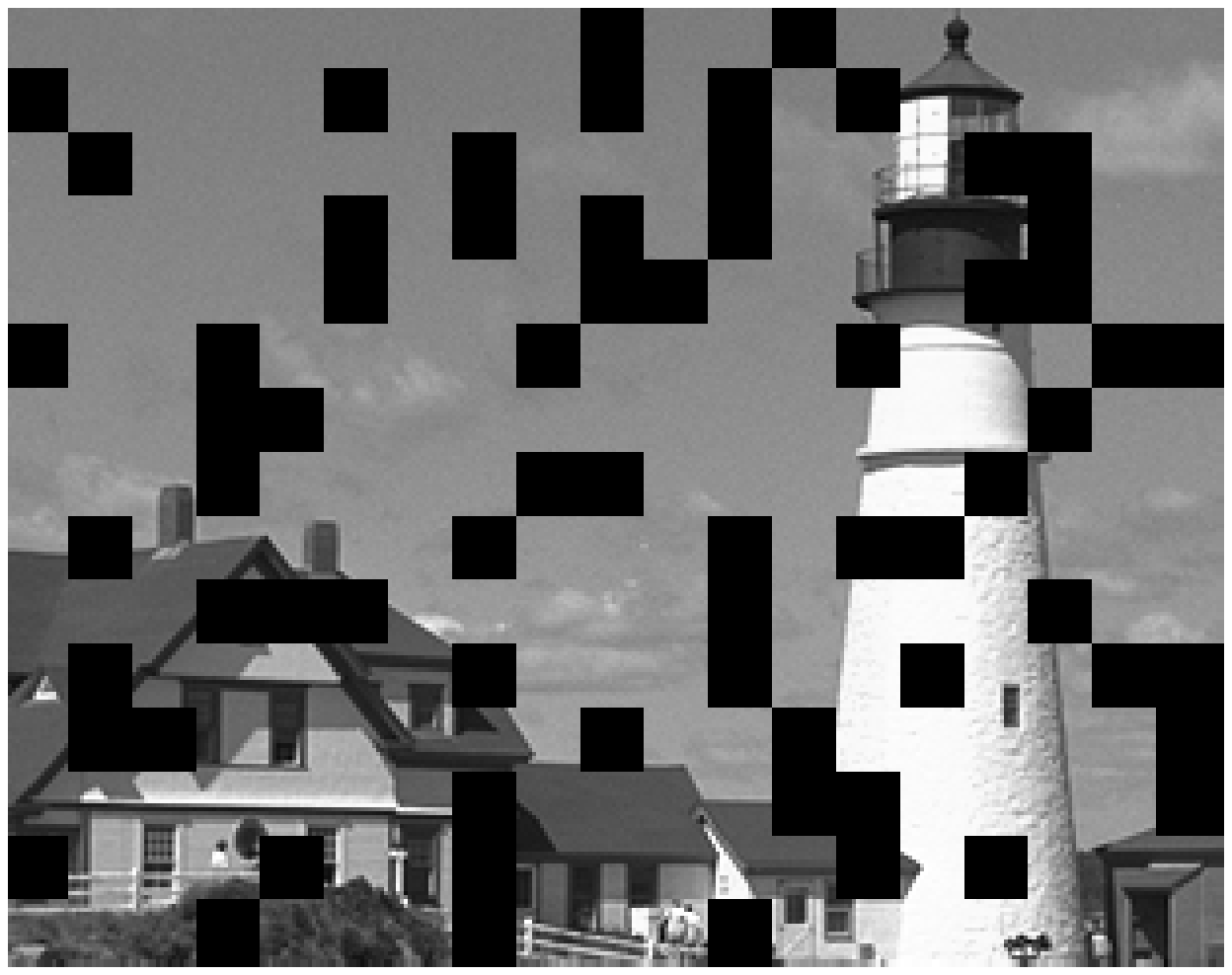}};
	\node (c) [right = -0.15cm of b] {\includegraphics[width=0.24\linewidth]{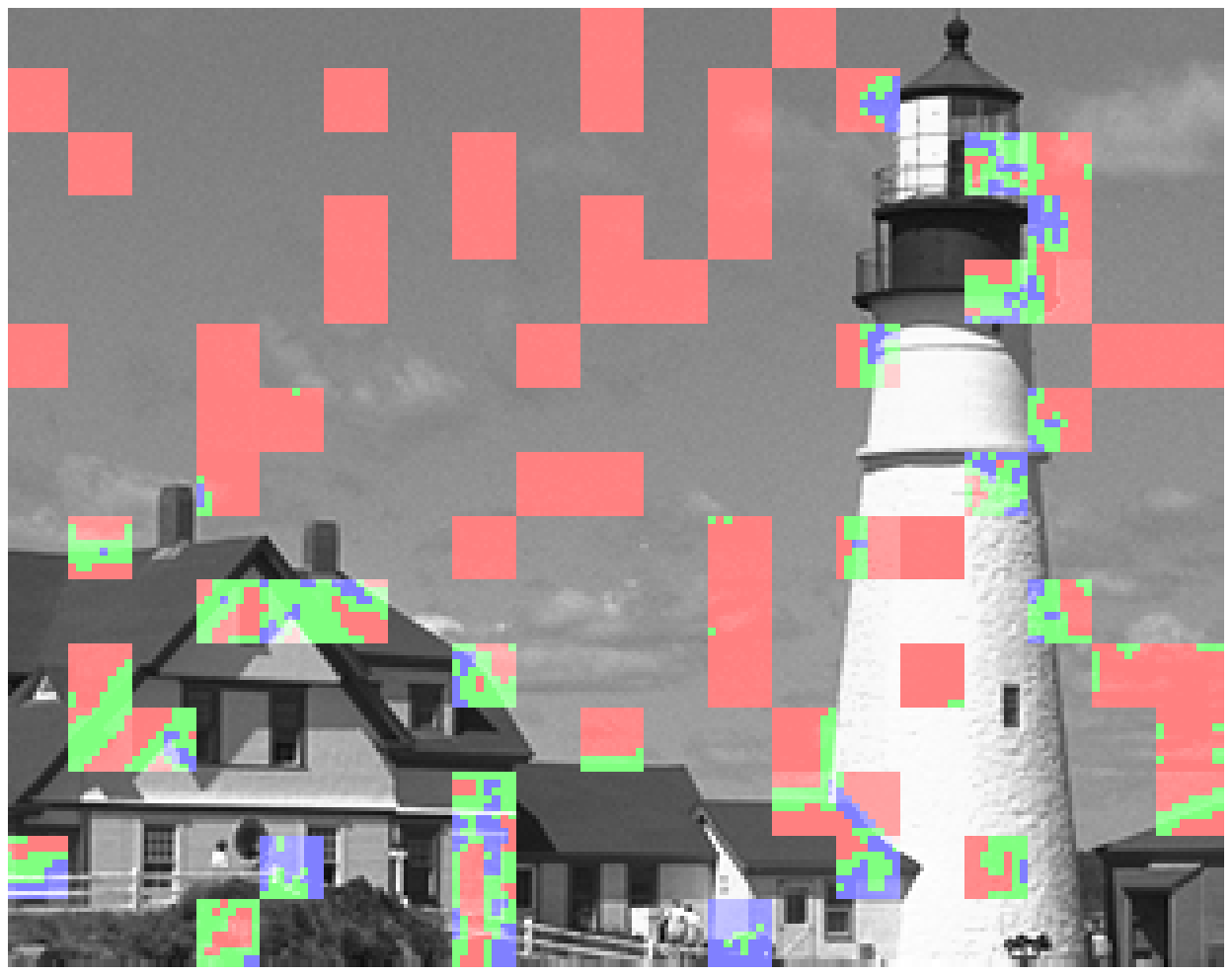}};
	\node (d) [right = -0.15cm of c] {\includegraphics[width=0.24\linewidth]{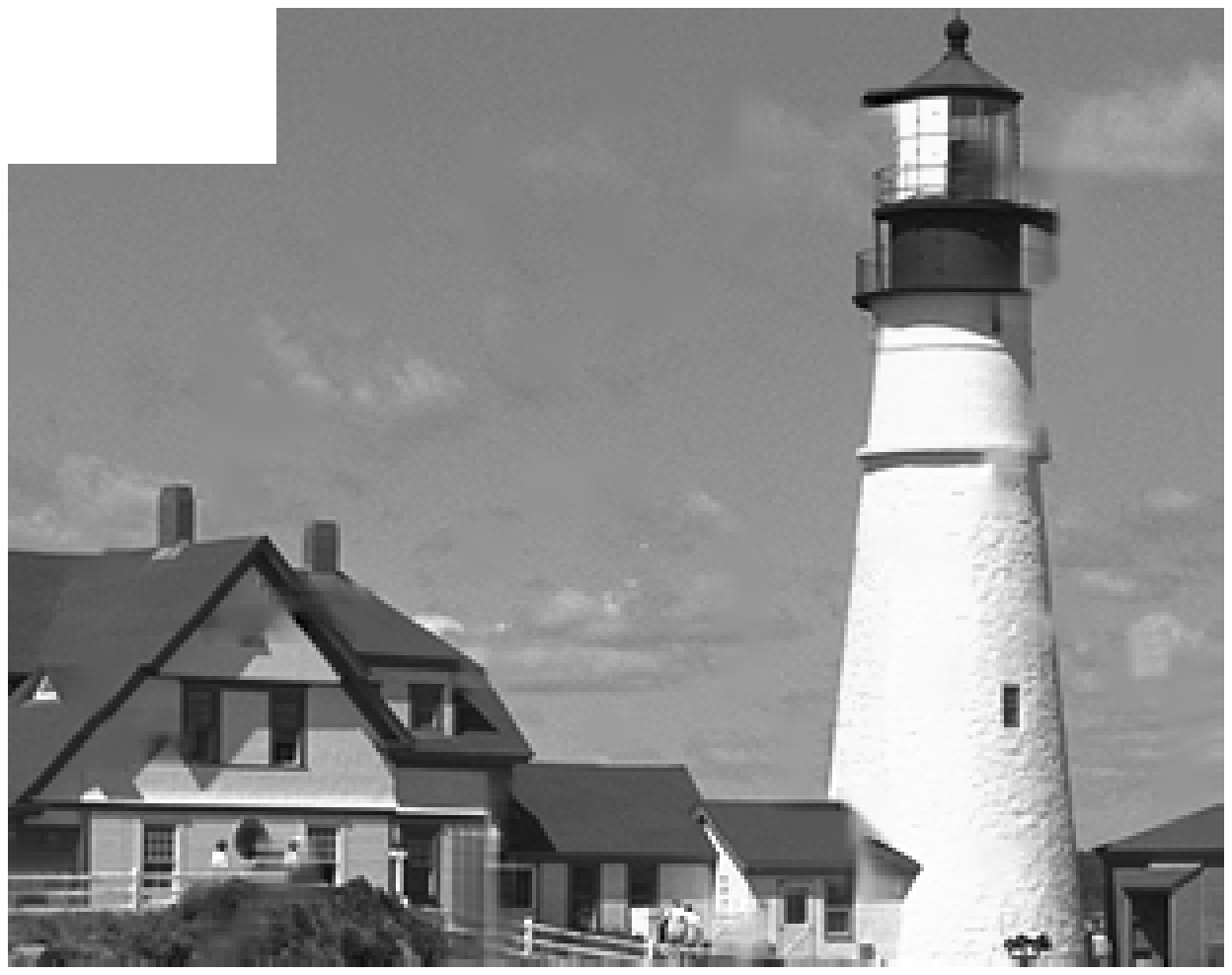}};
	
	\node (a_label) [below = -0.1cm of a] {\small{(a)}};
	\node (b_label) [below = -0.1cm of b] {\small{(b)}};
	\node (c_label) [below = -0.1cm of c] {\small{(c)}};
	\node (d_label) [below = -0.1cm of d] {\small{(d)}};
		
	\node (e) [below = 0.25cm of a] {\includegraphics[width=0.24\linewidth]{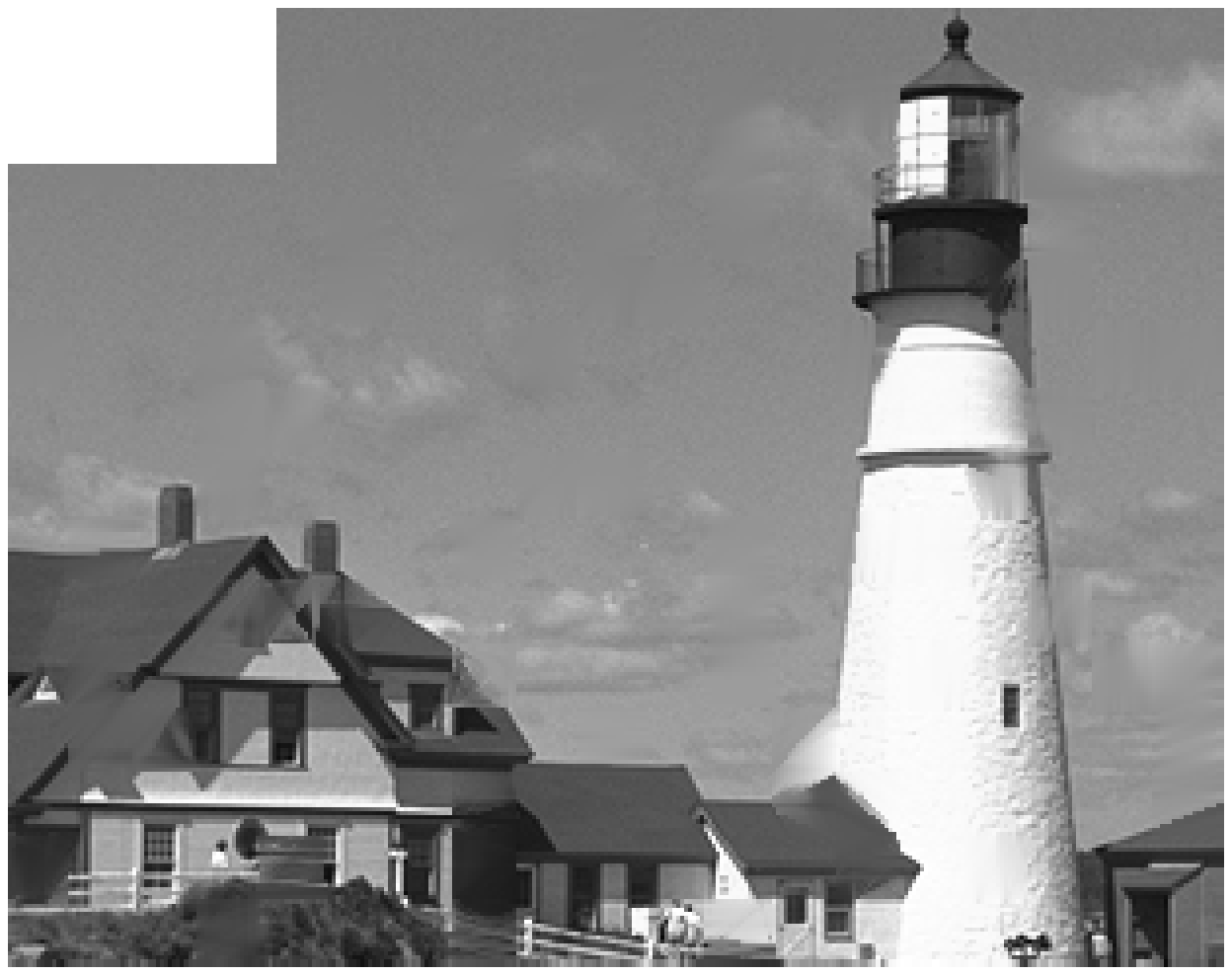}};
	\node (f) [right = -0.15cm of e] {\includegraphics[width=0.24\linewidth]{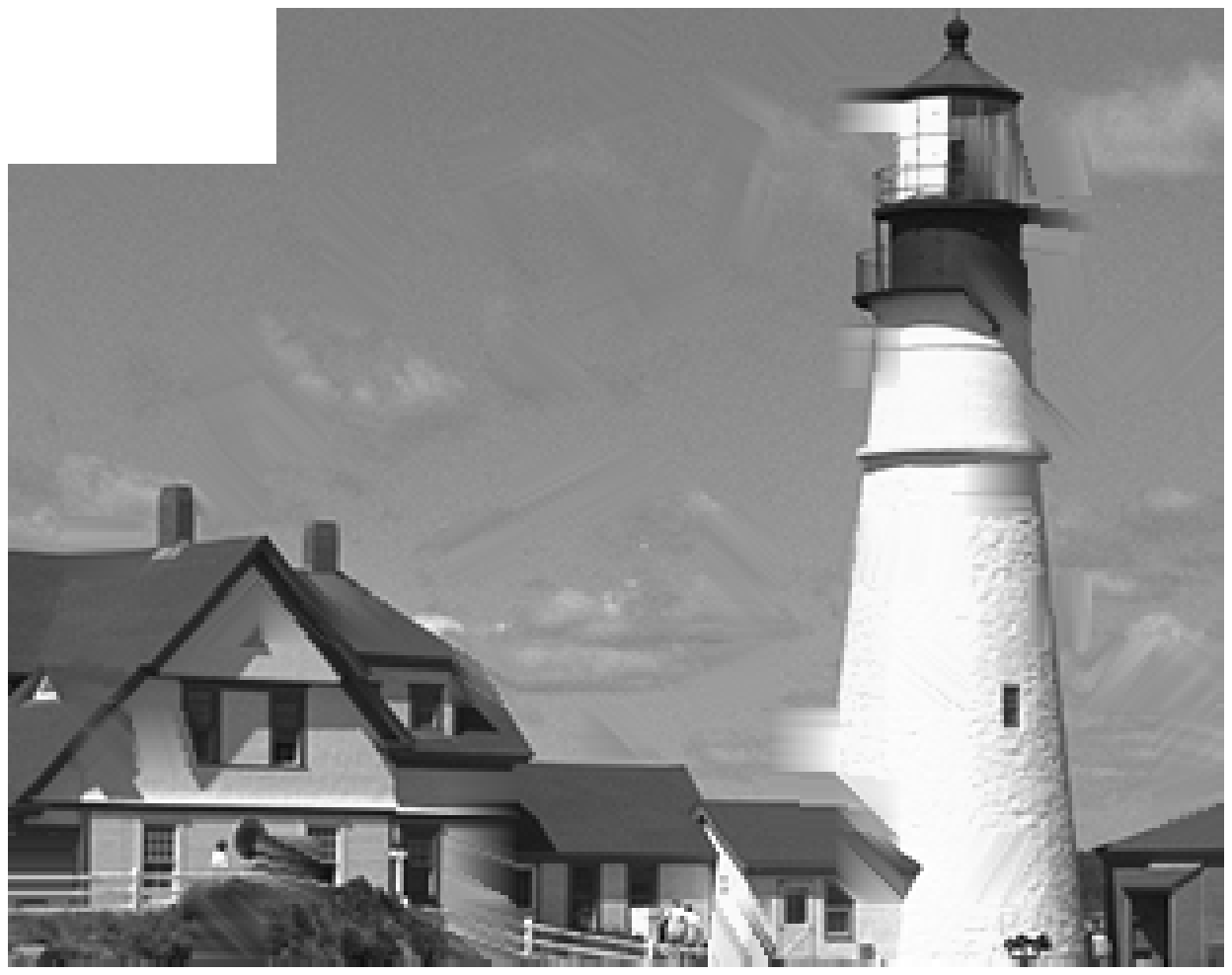}};
	\node (g) [right = -0.15cm of f] {\includegraphics[width=0.24\linewidth]{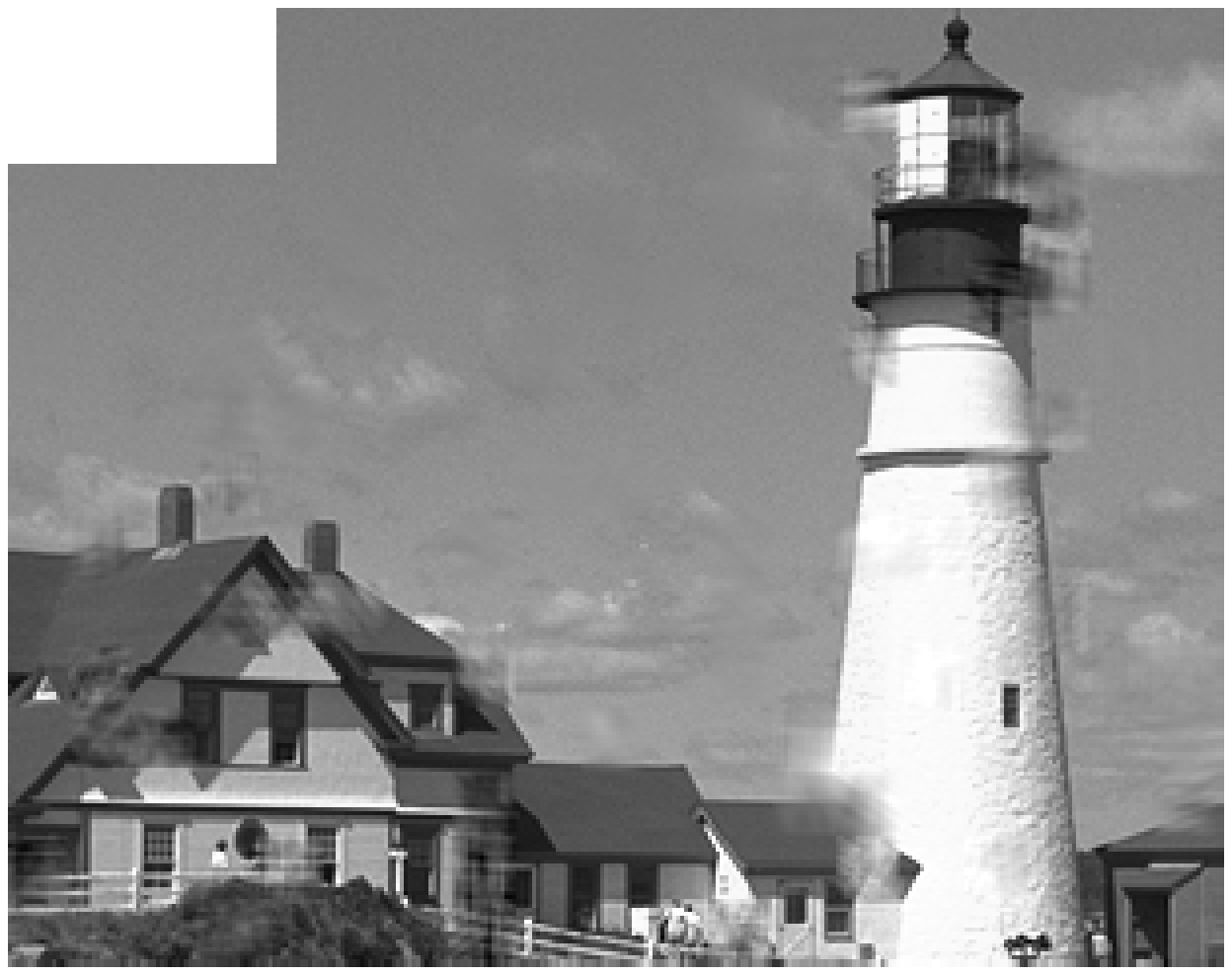}};
	\node (h) [right = -0.15cm of g] {\includegraphics[width=0.24\linewidth]{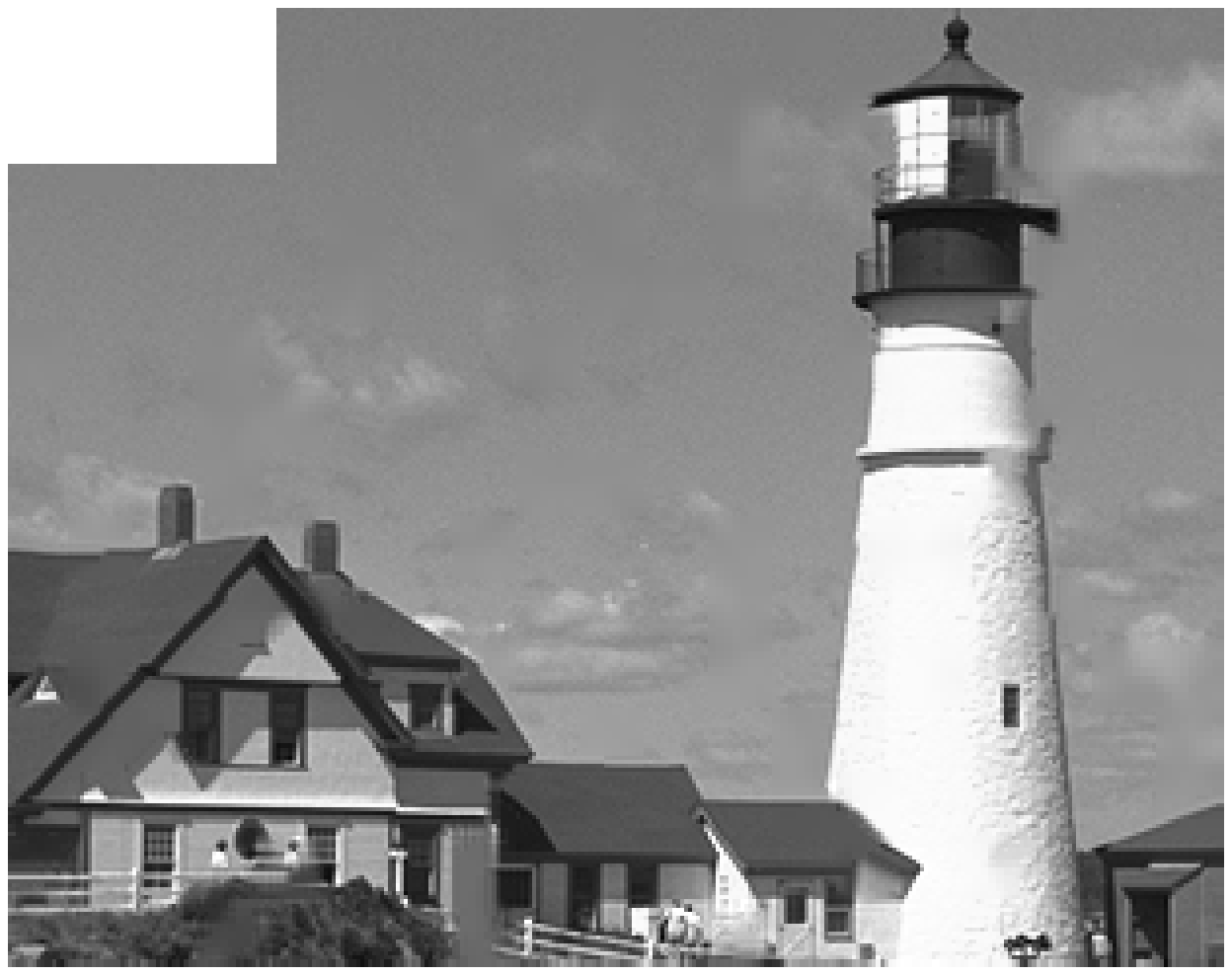}};
	
	\node (e_label) [below = -0.1cm of e] {\small{(e)}};
	\node (f_label) [below = -0.1cm of f] {\small{(f)}};
	\node (g_label) [below = -0.1cm of g] {\small{(g)}};
	\node (h_label) [below = -0.1cm of h] {\small{(h)}};
				
	\node (d_psnr) [above left = -0.5cm and -1.15cm of d] {\scriptsize{\textbf{27.84dB}}};
	\node (e_psnr) [above left = -0.5cm and -1.15cm of e] {\scriptsize{\textbf{25.21dB}}};
	\node (f_psnr) [above left = -0.5cm and -1.15cm of f] {\scriptsize{\textbf{26.71dB}}};
	\node (g_psnr) [above left = -0.5cm and -1.15cm of g] {\scriptsize{\textbf{27.31dB}}};
	\node (h_psnr) [above left = -0.5cm and -1.15cm of h] {\scriptsize{\textbf{27.80dB}}};
	
	\node (d_time) [above left = -0.75cm and -1.15cm of d] {\scriptsize{\textbf{79.65ms}}};
	\node (e_time) [above left = -0.75cm and -1.15cm of e] {\scriptsize{\textbf{13.98ms}}};
	\node (f_time) [above left = -0.75cm and -1.15cm of f] {\scriptsize{\textbf{9.29ms}}};
	\node (g_time) [above left = -0.75cm and -1.15cm of g] {\scriptsize{\textbf{66.26ms}}};
	\node (h_time) [above left = -0.75cm and -1.15cm of h] {\scriptsize{\textbf{12.14ms}}};
	
      \end{tikzpicture}
    \end{center}
    \vspace{-0.5cm}
    \caption{\small{Comparison of reconstructions obtained by different techniques and employing \textit{random} error pattern. (a) Original image from Kodak database, frame \#21. (b) Received frame. (c) Layer distribution for SK-MMSE efficient profile. Patches reconstructed by BRL are in red, by IDL in green and by HQL in blue. Reconstruction by (d) K-MMSE, (e) MDI, (f) EVC, (g) FSE, (h) SK-MMSE (efficient profile). The corresponding PSNR values (in dB) and computational times (in milliseconds per patch) are also indicated. Best to be viewed enlarged on a screen.}}
    \label{fig:example_random}  
\end{figure*}

Figures \ref{fig:example_dispersed} and \ref{fig:example_random} offer a subjective comparison for dispersed and random loss patterns, respectively. For illustration purposes, the efficient SK-MMSE reconstruction profile is used for both cases. Figures \ref{fig:example_dispersed}(c) and \ref{fig:example_random}(c) show the layer usage and illustrate how SK-MMSE automatically applies higher layers to more complex structures. It is also shown that \mbox{SK-MMSE} leads to a reconstruction quality virtually equivalent to \mbox{K-MMSE}. It can be observed that SK-MMSE (efficient profile) and K-MMSE produce the best result both on subjective and objective levels. However, the SK-MMSE estimator performs with considerable lower computational burden.

\vspace{0.25cm}

\subsection{Experiments with video sequences}
\label{subsec:video}

In order to analyse how the reconstruction quality affects error propagation, we have conducted tests on video sequences. In addition to PSNR, which measures objective distortions, we also use video quality metric (VQM) as defined in \cite{VQM}. VQM aims at approximating the subjective perception of video quality and it takes into account spatial as well as temporal variations of video signals\footnote{Note that unlike PSNR and SSIM, lower VQM values indicate better video quality.}. In order to test the performance on different frame resolutions, we have employed test video sequences from \textit{ClassC} (832$\times$480) and \textit{ClassB} (1920$\times$1080, full HD) \cite{Bossen2013}. The first 32 frames of each sequence are coded using H.264/AVC codec \cite{Recom} and the intra-frame refreshment interval is set to 12. Different quality of the encoded streams is achieved by setting the quantisation parameter (QP) value to 20, 30 and 40. Similar to the case of still images, both dispersed and random error patterns are employed and the errors are imprinted frame-wise. Moreover, since the concealment is applied at the decoder, the resulting video will be affected also by error propagation.
        
\begin{table*}[t]	
	\centering
	\scriptsize
	\begin{tabular}{p{\pLen}p{\pLen}|p{\pLen}p{\pLen}p{\pLen}|p{\pLen}p{\pLen}p{\pLen}|p{\pLen}p{\pLen}p{\pLen}|p{\pLen}p{\pLen}p{\pLen}|p{\pLen}p{\pLen}p{\pLen}|p{\pLen}p{\pLen}p{\pLen}}
		\hline\hline
		& & \multicolumn{3}{c}{SLP-E} & \multicolumn{3}{c}{FSE} & \multicolumn{3}{c}{SK-MMSE$_{\text{xp}}$} & \multicolumn{3}{c}{SK-MMSE$_{\text{ff}}$} & \multicolumn{3}{c}{SK-MMSE$_{\text{xc}}$} & \multicolumn{3}{c}{K-MMSE} \\
		& & PSNR & VQM & Time & PSNR & VQM & Time & PSNR & VQM & Time & PSNR & VQM & Time & PSNR & VQM & Time & PSNR & VQM & Time \\ \hline
		\multirow{3}{*}{\textit{ClassB}}
		& QP$_{20}$ & 31.57 & 0.316 & 45.62 & 31.96 & 0.318 & 83.19 & 31.72 & 0.315 & 3.32 & 31.82 & 0.313 & 4.48 & 32.08 & 0.311 & 24.87 & 32.37 & 0.312 & 100.00\\
		& QP$_{30}$ & 30.39 & 0.317 & 45.62 & 30.82 & 0.326 & 83.19 & 30.54 & 0.316 & 3.10 & 30.64 & 0.315 & 4.14 & 30.64 & 0.314 & 22.72 & 31.03 & 0.316 & 100.00\\
		& QP$_{40}$ & 28.76 & 0.362 & 45.62 & 29.06 & 0.372 & 83.19 & 28.88 & 0.361 & 2.73 & 28.92 & 0.361 & 3.62 & 28.92 & 0.360 & 17.93 & 29.07 & 0.362 & 100.00\\ \hline
		
		\multirow{3}{*}{\textit{ClassC}} 
		& QP$_{20}$ & 29.05 & 0.559 & 45.62 & 28.86 & 0.617 & 83.19 & 29.32 & 0.555 & 8.05 & 29.33 & 0.547 & 11.48 & 29.43 & 0.540 & 45.72 & 29.47 & 0.526 & 100.00\\
		& QP$_{30}$ & 27.47 & 0.564 & 45.62 & 27.31 & 0.629 & 83.19 & 27.60 & 0.560 & 7.21 & 27.68 & 0.552 & 10.42 & 27.75 & 0.545 & 43.23 & 27.78 & 0.533 & 100.00\\
		& QP$_{40}$ & 25.38 & 0.613 & 45.62 & 25.44 & 0.696 & 83.19 & 25.50 & 0.611 & 5.77 & 25.53 & 0.605 & 8.37  & 25.53 & 0.600 & 37.08 & 25.57 & 0.593 & 100.00\\ \hline \hline
	\end{tabular}
	\caption{\small{Average PSNR (in dB), VQM, and relative computational time (in \%) with respect to full-featured K-MMSE for different SEC techniques. Video sequences of $\textit{ClassB}$ and $\textit{ClassC}$ are used and encoded with different QP values. \textit{Dispersed} error pattern is employed.}}
	\label{tab:video_dispersed}
\end{table*}

\begin{table*}[t]
	\centering
	\scriptsize
	\begin{tabular}{p{\pLen}p{\pLen}|p{\pLen}p{\pLen}p{\pLen}|p{\pLen}p{\pLen}p{\pLen}|p{\pLen}p{\pLen}p{\pLen}|p{\pLen}p{\pLen}p{\pLen}|p{\pLen}p{\pLen}p{\pLen}|p{\pLen}p{\pLen}p{\pLen}}
		\hline\hline
		& & \multicolumn{3}{c}{SLP-E} & \multicolumn{3}{c}{FSE} & \multicolumn{3}{c}{SK-MMSE$_{\text{xp}}$} & \multicolumn{3}{c}{SK-MMSE$_{\text{ff}}$} & \multicolumn{3}{c}{SK-MMSE$_{\text{xc}}$} & \multicolumn{3}{c}{K-MMSE} \\
		& & PSNR & VQM & Time & PSNR & VQM & Time & PSNR & VQM & Time & PSNR & VQM & Time & PSNR & VQM & Time & PSNR & VQM & Time \\ \hline
		\multirow{3}{*}{\textit{ClassB}}
		& QP$_{20}$ & 30.39 & 0.378 & 45.62 & 30.59 & 0.393 & 83.19 & 30.45 & 0.378 & 2.91 & 30.53 & 0.375 & 3.89 & 30.66 & 0.374 & 21.70 & 30.63 & 0.372 & 100.00\\
		& QP$_{30}$ & 29.29 & 0.378 & 45.62 & 29.59 & 0.395 & 83.19 & 29.38 & 0.377 & 2.75 & 29.46 & 0.376 & 3.63 & 29.54 & 0.375 & 19.94 & 29.54 & 0.374 & 100.00\\
		& QP$_{40}$ & 27.94 & 0.414 & 45.62 & 28.20 & 0.431 & 83.19 & 28.02 & 0.414 & 2.46 & 28.08 & 0.412 & 3.18 & 28.09 & 0.412 & 15.66 & 28.06 & 0.411 & 100.00\\ \hline
		
		\multirow{3}{*}{\textit{ClassC}}
		& QP$_{20}$ & 27.54 & 0.655 & 45.62 & 27.64 & 0.761 & 83.19 & 27.75 & 0.650 & 6.47 & 27.84 & 0.642 & 9.36 & 27.91 & 0.634 & 41.09 & 27.92 & 0.619 & 100.00\\
		& QP$_{30}$ & 26.11 & 0.656 & 45.62 & 26.20 & 0.775 & 83.19 & 26.27 & 0.652 & 5.85 & 26.33 & 0.645 & 8.50 & 26.33 & 0.639 & 39.32 & 26.33 & 0.621 & 100.00\\
		& QP$_{40}$ & 24.44 & 0.686 & 45.62 & 24.61 & 0.834 & 83.19 & 24.50 & 0.684 & 4.70 & 24.55 & 0.676 & 6.70 & 24.58 & 0.671 & 33.12 & 24.56 & 0.659 & 100.00\\ \hline \hline
	\end{tabular}
	\caption{\small{Average PSNR (in dB), VQM, and relative computational time (in \%) with respect to full-featured K-MMSE for different SEC techniques. Video sequences of $\textit{ClassB}$ and $\textit{ClassC}$ are used and encoded with different QP values. \textit{Random} error pattern is employed.}}
	\label{tab:video_random}
\end{table*}

Tables \ref{tab:video_dispersed} and \ref{tab:video_random} show the average PSNR and VQM values for dispersed and random error pattern, respectively. The average computational time for different EC techniques is also indicated and the results are provided for the three different values of QP. Note that K-MMSE, SLP-E and FSE are not content dependent and therefore their computational times are constant.

It can be observed that for \textit{ClassC} the excellent profile yields virtually the same quality as the full-featured K-MMSE while requiring less than a half of its computational time. The express profile is around 15 times faster and presents only a moderate loss of approximately 0.1dB (in terms of PSNR) with respect to the original K-MMSE approach. In addition, the proposed technique outperforms other state-of-the-art EC methods both in terms of PSNR and VQM while exhibiting considerably lower computational complexity (depending on the selected profile).

Note that the size of a macroblock is constant regardless of the video resolution. Therefore, one macroblock from high resolution video stream contains, in general, less structure than its lower-resolution counterpart. Therefore, SK-MMSE will tend to exploit more lower (and faster) reconstruction layers. This behaviour can be observed by analysing the results obtained for \textit{ClassB}. First, note that SK-MMSE is on average more than 2 times faster with respect to the results of \textit{ClassC} and the express profile can exhibit a computational burden more than 40 times lower than the original K-MMSE approach. Second, it can be seen that for some cases FSE exhibits slightly better PSNR behaviour than some of the proposed SK-MMSE profiles. However, FSE involves up to 27 times higher computational burden and, in addition, the proposed SK-MMSE method (including all its profiles) consistently outperforms both FSE and SLP-E in terms of VQM.

Moreover, it is shown that the proposed method achieves higher speed-up with increasing QP values. This is due to the fact that higher QP values yield visually smoother encoded streams where fine structures are not present. Therefore, more complex layers, i.e. IDL and HQL, are less frequently required.

Finally, note that the proposed SK-MMSE algorithm is susceptible to multi-level parallelisation. In other words, the reconstruction process can be massively parallelised on both block-based and patch-based levels. This is, however, not a straightforward task since inter-block and inter-patch dependencies have to be taken into account in order to avoid multiple blocks/patches being processed in parallel. Even though this multi-level parallelisation can yield significant computational time reductions it must be noted that, unlike the proposed scalable scheme, parallelisation does not reduce the \mbox{computational burden}.

\section{Conclusions}
\label{sec:conclusions}
We have proposed a new scalable error concealment technique based on \mbox{K-MMSE} that aims at reducing the processing time while preserving the high reconstruction quality of \mbox{K-MMSE}. We have introduced three reconstruction layers that emanate from the original \mbox{K-MMSE} framework. These layers are hierarchically stacked so lower layers feed information to higher layers. In general, higher reconstruction layers produce more accurate reconstructions at the expense of higher computational complexity. However, the use of high layers in homogeneous or stationary image areas is a waste of resources since reconstruction of virtually the same quality can be obtained by lower layers. Therefore, we have designed a layer selection mechanism based on profiles. By analysing the visual properties of the available surrounding area, the profile controls which is the highest layer to be applied during the reconstruction of the lost region. Thus, by adjusting the profiles, the trade-off between speed and quality can be efficiently controlled so SK-MMSE can be adapted to different platforms, distortion rates or available computational power. It has been shown that our proposal can achieve a speed-up of up to 10 times with respect to K-MMSE with negligible loss in reconstruction quality. Simulations reveal that our proposal outperforms other-state-of-the-art techniques in terms of quality and required computational time.

Ongoing work is focused on improving the proposed method by introducing a multi-level parallelisation scheme.

\bibliographystyle{IEEEtran}
\bibliography{references}

\begin{thebibliography}{10}
\providecommand{\url}[1]{#1}
\csname url@samestyle\endcsname
\providecommand{\newblock}{\relax}
\providecommand{\bibinfo}[2]{#2}
\providecommand{\BIBentrySTDinterwordspacing}{\spaceskip=0pt\relax}
\providecommand{\BIBentryALTinterwordstretchfactor}{4}
\providecommand{\BIBentryALTinterwordspacing}{\spaceskip=\fontdimen2\font plus
\BIBentryALTinterwordstretchfactor\fontdimen3\font minus
  \fontdimen4\font\relax}
\providecommand{\BIBforeignlanguage}[2]{{%
\expandafter\ifx\csname l@#1\endcsname\relax
\typeout{** WARNING: IEEEtran.bst: No hyphenation pattern has been}%
\typeout{** loaded for the language `#1'. Using the pattern for}%
\typeout{** the default language instead.}%
\else
\language=\csname l@#1\endcsname
\fi
#2}}
\providecommand{\BIBdecl}{\relax}
\BIBdecl

\bibitem{ISEC_multibroadcast}
{T. Tr\"{o}ger and A. Kaup}, ``Inter-sequence error concealment techniques for
  multi-broadcast {TV} reception,'' \emph{IEEE Transactions on Broadcasting},
  vol.~57, pp. 777--793, October 2011.

\bibitem{Sandvine}
\url{https://www.sandvine.com/trends/global-internet-phenomena/}.

\bibitem{Richardson2010}
I.~Richardson, ``The {H}.264 advanced video compression standard,'' Wiley,
  2010.

\bibitem{HEVC_overview}
G.~Sullivan, J.-R. Ohm, W.-J. Han, and T.~Wiegand, ``Overview of the high
  efficiency video coding ({HEVC}) standard,'' \emph{IEEE Transactions on
  Circuits and Systems for Video Technology}, vol.~22, pp. 1649--1668, December
  2012.

\bibitem{EC_DTV}
J.-W. Suh and Y.-S. Ho, ``Error concealment techniques for digital {TV},''
  \emph{IEEE Transactions on Broadcasting}, vol.~48, pp. 299--306, December
  2002.

\bibitem{HEVC_syntax}
R.~Sj{\"{o}}berg, Y.~Chen, A.~Fujibayashi, M.~Hannuksela, J.~Samuelsson,
  T.~Tan, Y.-K. Wang, and S.~Wenger, ``Overview of {HEVC} high-level syntax and
  reference picture management,'' \emph{IEEE Transactions on Circuits and
  Systems for Video Technology}, vol.~22, pp. 1858--1870, December 2012.

\bibitem{FadingEC}
M.~Friebe and A.~Kaup, ``Fading techniques for error concealment in block-based
  video coding systems,'' \emph{IEEE Transactions on Broadcasting}, vol.~53,
  pp. 286--296, March 2007.

\bibitem{FEC_Peinado}
{A.M. G\'{o}mez}, {A.M. Peinado}, {V. S\'{a}nchez}, and {A.J. Rubio},
  ``Combining media specific {F}{E}{C} and error concealment for robust
  distributed speech recognition over loss-prone packet channels,'' \emph{IEEE
  Transactions on Multimedia}, vol.~8, pp. 1228--1238, November 2006.

\bibitem{Hybrid_EC}
M.~Hwang, J.~Kim, D.~Duong, and S.~Ko, ``Hybrid temporal error concealment
  methods for block-based compressed video transmission,'' \emph{IEEE
  Transactions on Broadcasting}, vol.~54, pp. 198--207, June 2008.

\bibitem{AVC}
T.~Wiegand, G.~Sullivan, G.~Bjontegaard, and A.~Luthra, ``Overview of the
  {H}.264/{A}{V}{C} video coding standard,'' \emph{IEEE Transactions on
  Circuits and Systems for Video Technology}, vol.~13, pp. 560--576, July 2003.

\bibitem{BLI}
P.~Salama, N.~Shroff, E.~Coyle, and E.~Delp, ``Error concealment techniques for
  encoded video streams,'' in \emph{Proceedings of ICIP}, 1995, pp. 9--12.

\bibitem{EXT}
Y.~Zhao, H.~Chen, X.~Chi, and J.~Jin, ``Spatial error concealment using
  directional extrapolation,'' in \emph{Proceedings of DICTA}, 2005, pp.
  278--283.

\bibitem{Robie}
D.~Robie and R.~Mersereau, ``The use of {H}ough transforms in spatial error
  concealment,'' in \emph{Proceedings of ICASSP}, vol.~4, 2000, pp. 2131--2134.

\bibitem{Gharavi}
H.~Gharavi and S.~Gao, ``Spatial interpolation algorithm for error
  concealment,'' in \emph{Proceedings of ICASSP}, April 2008, pp. 1153--1156.

\bibitem{MRF}
S.~Shirani, F.~Kossentini, and R.~Ward, ``An adaptive {M}arkov random field
  based error concealment method for video communication in error prone
  environment,'' in \emph{Proceedings of ICIP}, vol.~6, 1999, pp. 3117--3120.

\bibitem{Harrison}
P.~Harrison, ``Texture synthesis, texture transfer and plausible restoration,''
  Ph.D. dissertation, Monash University, 2005.

\bibitem{Criminisi}
{A. Criminisi and P. P\'{e}rez and K. Toyama}, ``Region filling and object
  removal by exemplar-based image inpainting,'' \emph{IEEE Transactions on
  Image Processing}, vol.~13, pp. 1200--1212, September 2004.

\bibitem{OAI}
X.~Li and M.~Orchard, ``Novel sequential error-concealment techniques using
  orientation adaptive interpolation,'' \emph{IEEE Transactions on Circuits and
  Systems for Video Technology}, vol.~12, pp. 857--864, October 2002.

\bibitem{EVC}
J.~Koloda, V.~S\'{a}nchez, and A.~M. Peinado, ``Spatial error concealment based
  on edge visual clearness for image/video communication,'' \emph{Circuits,
  Systems and Signal Processing}, April 2013.

\bibitem{BayesianPyramid}
G.~Zhai, X.~Yang, W.~Lin, and W.~Zhang, ``Bayesian error concealment with
  {D}{C}{T} pyramid for images,'' \emph{IEEE Transactions on Circuits and
  Systems for Video Technology}, vol.~20, pp. 1224--1232, September 2010.

\bibitem{BLF}
G.~Zhai, J.~Cai, W.~Lin, X.~Yang, and W.~Zhang, ``Image error-concealment via
  block-based bilateral filtering,'' \emph{IEEE International Conference on
  Multimedia and Expo}, pp. 621--624, June 2008.

\bibitem{SLPE}
J.~Koloda, J.~\O{}stergaard, S.~Jensen, V.~S\'{a}nchez, and A.~Peinado,
  ``Sequential error concealment for video/images by sparse linear
  prediction,'' \emph{IEEE Transactions on Multimedia}, vol.~15, pp. 957--969,
  June 2013.

\bibitem{Liu2015}
J.~Liu, G.~Zhai, X.~Tang, B.~Yang, and L.~Chen, ``Spatial error concealment
  with an adaptive linear predictor,'' \emph{IEEE Transactions on Circuits and
  Systems for Video Technology}, vol.~25, pp. 353--366, March 2015.

\bibitem{POCS}
H.~Sun and W.~Kwok, ``Concealment of damaged block transform coded images using
  projections onto convex sets,'' \emph{IEEE Transactions on Image Processing},
  vol.~4, no.~4, April 1995.

\bibitem{CVFSE}
{J. Seiler and A. Kaup}, ``Complex-valued frequency selective extrapolation for
  fast image and video signal extrapolation,'' \emph{IEEE Signal Processing
  Letters}, vol.~17, pp. 949--952, November 2010.

\bibitem{FOFSE}
J.~Seiler and A.~Kaup, ``Fast orthogonality deficiency compensation for
  improved frequency selective image extrapolation,'' in \emph{Proceedings of
  ICASSP}, April 2008, pp. 781--784.

\bibitem{MDI}
H.~Asheri, H.~Rabiee, N.~Pourdamghani, and M.~Ghanbari, ``Multi-directional
  spatial error cocnealment using adaptive edge thresholding,'' \emph{IEEE
  Transactions on Consumer Electronics}, vol.~58, pp. 880--885, 2012.

\bibitem{Efficient_STEC}
G.-L. Wu, C.-Y. Chen, T.-H. Wu, and S.-Y. Chien, ``Efficient spatial-temporal
  error concealment algorithm and hardware architecture design for h.264/avc,''
  \emph{IEEE Transactions on Circuits and Systems for Video Technology},
  vol.~20, pp. 1409--1422, November 2010.

\bibitem{FE_SEC}
N.~Beheiry, M.~Sharkawy, M.~Lofty, and S.~Elnoubi, ``An adaptive fast and
  efficient spatial error concealment technique for block-based video coding
  systems,'' in \emph{Proceedings of MWSCAS}, 2009, pp. 663--668.

\bibitem{CoopGame}
C.-L. Ho and L.-W. Chang, ``Temporal and spatial error concealment using
  cooperative game,'' in \emph{Proceedings of ISEEE}, 2014, pp. 380--384.

\bibitem{KungKimKuo}
W.~Kung, C.~Kim, and C.~Kuo, ``Spatial and temporal error concealment
  techniques for video transmission over noisy channels,'' \emph{IEEE
  Transactions on Circuits and Systems for Video Technology}, vol.~16, pp.
  789--802, July 2006.

\bibitem{Hsia2016}
S.-C. Hsia and C.~Hsiao, ``Fast-efficient shape error concealment technique
  based on block classification,'' \emph{IET Image Processing}, vol.~10, pp.
  693--700, October 2016.

\bibitem{CAD}
Z.~Rongfu, Z.~Yuanhua, and H.~Xiaodong, ``Content-adaptive spatial error
  concealment for video communication,'' \emph{IEEE Transactions on Consumer
  Electronics}, vol.~50, pp. 335--341, February 2004.

\bibitem{KMMSE}
J.~Koloda, A.~Peinado, and V.~S\'{a}nchez, ``Kernel-based {MMSE} multimedia
  signal reconstruction and its application to spatial error concealment,''
  \emph{IEEE Transactions on Multimedia}, vol.~16, pp. 1729--1738, October
  2014.

\bibitem{Tecnick_v3}
N.~Asuni and A.~Giachetti, ``{TESTIMAGES}: {A} large-scale archive for testing
  visual devices and basic image processing algorithms,'' in \emph{Proceedings
  of STAG}, 2014.

\bibitem{SSIM}
Z.~Wang, A.~Bovik, H.~Sheikh, and E.~Simoncelli, ``Image quality assessment:
  from error visibility to structural similarity,'' \emph{IEEE Transactions on
  Image Processing}, vol.~13, pp. 600--612, April 2004.

\bibitem{Kodak}
``Kodak image dataset,'' \url{http://r0k.us/graphics/kodak/}, {A}ccessed:
  29-03-2015.

\bibitem{VQM}
S.~Li, L.~Ma, and K.~Ngan, ``Full-reference video quality assessment by
  decoupling detail losses and additive impairments,'' \emph{IEEE Transactions
  on Circuits and Systems for Video Technology}, vol.~22, pp. 1100--1112, July
  2012.

\bibitem{Bossen2013}
F.~Bossen, ``Common test conditions and software reference configurations,''
  document JCTVC-L1100, ITU-T VCEG and ISO/IEC MPEG (JCT-VC), Geneva,
  Switzerland, January 2013.

\bibitem{Recom}
{I}{T}{U}-{T}, ``{I}{T}{U}-{T} {R}ecommendation {H}.264,'' {I}nternational
  {T}elecommunication {U}nion, 2010.

\end{thebibliography}

\end{document}